\documentclass[a4,center,fleqn,10pt,twocolumn]{article}
\usepackage{amsmath, amssymb}
\usepackage{graphics,url}
\usepackage{crop,inputenc,amsmath,amssymb,latexsym,float,epsfig,wrapfig,graphicx,stfloats, multicol,rotating,multirow,dcolumn, boxedminipage,makeidx,soul,url,xspace,times} 
\DeclareMathOperator{\dc}{Ca}
\newcounter{definition}
\newenvironment{definition}[1][]{\refstepcounter{definition}\par\medskip\noindent%
\textbf{Definition~\thedefinition. #1} \rmfamily}{\medskip}
\newcommand{\ignore}[1]{}
\renewcommand\baselinestretch{1}
\newcommand{\authors}{
\begin{tabular}{c}
Giuseppe Jurman${}^1$, Samantha Riccadonna${}^1$,\\
Roberto Visintainer${}^{1,2}$ and Cesare Furlanello${}^1$\\
\\
${}^1$Fondazione Bruno Kessler\\
I-38123 Povo (Trento), Italy\\
${}^2$DISI, University of Trento\\
I-38123 Povo (Trento), Italy\\

\end{tabular}
\\
\small{\texttt{\{jurman,riccadonna,visintainer,furlan\}@fbk.eu}}
}

\begin{document}
\title{Algebraic Comparison of Partial Lists in Bioinformatics}
\author{\authors}
\date{\today}

\maketitle

\begin{abstract} 
The outcome of a functional genomics pipeline is usually a partial list of genomic features, ranked by their relevance in modelling biological phenotype in terms of a classification or regression model. 
Due to resampling protocols or just within a meta-analysis comparison, instead of one list it is often the case that sets of alternative feature lists (possibly of different lengths) are obtained. 
Here we introduce a method, based on the algebraic theory of symmetric groups, for studying the variability between lists (``list stability'') in the case of lists of unequal length. 
We provide algorithms evaluating stability for lists embedded in the full feature set or just limited to the features occurring in the partial lists.  
The method is demonstrated first on synthetic data in a gene filtering task and then for finding gene profiles on a recent prostate cancer dataset.  
\end{abstract}

\section{INTRODUCTION}
Defining indicators for assessing ranked lists variability has become a key research issue in functional genomics \cite{boulesteix09stability,ein-dor06thousands,boutros09prognostic,lau07three-gene, shi09tale}.

In \cite{jurman08algebraic}, a method is introduced to detect stability (homogeneity) of a set of ranked lists of biomarkers emerging as output of feature selection algorithm during a molecular profiling task.  
The stability indicator relies on the application of metric methods for ordered data viewed as elements of a suitable permutation group: foundations of such theory can be found in \cite{critchlow85metric,diaconis88group} and it is based on the concept of distance between two lists; in particular, the employed metric is the Canberra distance as discussed in \cite{lance67mixed}.  
The mathematical details of the stability procedure are described in \cite{jurman09canberra,gobbi08algebraic}: given a set of ordered lists, the basic mechanism is to evaluate the degree of self-homogeneity of a set of ordered lists through the computation of all the mutual distances between the elements of the original set.  
Moreover, by using the location parameter, the Canberra distance can also be computed between upper partial lists of the original complete lists, the so called top-$k$ lists \cite{fagin03comparing}, formed by their $k$ best ranked elements. 
The method proposed in \cite{jurman08algebraic} has a main drawback limiting its application in many practical situations: the studied lists are required to have the same length.  
As complete lists, they all share the same elements, with only their ordering being different and, when considering partial top-$k$ lists, the same $k$ must be chosen for all sublists.  

This is usually not the case when investigating the outcomes of profiling experiments where the employed feature ranking method does not produce a rank for every involved feature, but it just scores the best performing ones, thus leading to the construction of lists with different lengths.
Some work towards partial lists comparison has recently appeared in literature, both for general contexts \cite{bar-ilan06methods} and more focussed on the gene ranking case \cite{fury06overlapping,pearson07reciprocal,yang07meta-analysis}, but they all consist of set-theoretical measures.

In the present work we propose an extension of the method introduced in \cite{jurman08algebraic} to mend this flaw by allowing computing (Canberra) distance for two lists of different length, defined within the framework of the metric methods for permutation groups.  
The novel approach is based on the use of quotients of permutation groups. 
The key formula can be split into two main components: one taking care of the elements occurring in the selected lists, and the second one considering the remaining elements of the complete set of features the experiment started from. 
In particular, this second component is independent from the positions of the selected elements in the lists: neglecting this part, a different stability measure (called the core component of the complete formula) is obtained.
An applications of the described methods can be found in \cite{maqc09maqcII}.

After having detailed the central algorithm in Section \ref{sec:math}, applications to synthetic and genomics datasets and different machine learning tasks are discussed in Section \ref{sec:results}.
The described algorithm is publicly available within the Python package \texttt{mlpy} (\url{https://mlpy.fbk.eu}) for statistical machine learning.

\section{MATERIALS AND METHODS}
\label{sec:math}
\subsection{Introduction}
The procedure described in \cite{jurman08algebraic} is made of two
separate parts, the former concerning the computation of all the
mutual distances between the (complete or partial) lists, and the
latter the construction of the matrix starting from those distances
and of the indicator coming from the defined matrix.  This
second phase is independent from the length of the considered lists:
the extension shown hereafter only affects the previous step,
\textit{i.e.} the distance definition.
The algorithm relies on application of metric methods
for ordered data viewed as elements of a suitable permutation group:
foundations of such theory can be found in
\cite{kendall62rank,diaconis77spearman,critchlow85metric,diaconis88group} and it is based on the
concept of distance between two lists; in particular, the employed
metric is the Canberra distance \cite{lance67mixed}.  The
mathematical details of the procedure are shown in
\cite{jurman09canberra,gobbi08algebraic}.  
The described algorithm defines a measure to
evaluate the degree of self-homogeneity of a set of ordered lists
through the computation of all the mutual distances between the
elements of the original set.  Moreover, the distance can also be
computed between upper partial lists of the original complete lists,
the so called top-$k$ lists \cite{fagin03comparing}, formed by their
$k$ best ranked elements.
The indicator introduced in \cite{jurman08algebraic} has a main drawback limiting its application in
many practical situations: the studied lists are required to have the
same length.  In fact, as complete lists they must share the same
elements, with only their ordering being different and, when
considering partial top-$k$ lists, the same $k$ must be chosen for all
sublists.  This is usually not the case when investigating the
outcomes of profiling experiments where the employed feature ranking
method does not produce a rank for every involved feature, but it just
scores the best performing ones, thus leading to the construction of
lists with different length.
Some work towards this task has recently appeared in literature, both
for general contexts \cite{bar-ilan06methods} and more focussed on the
gene ranking case
\cite{fury06overlapping,pearson07reciprocal,yang07meta-analysis}, but
they all consist of set-theoretical measure.
In the present work we propose an extension of the method introduced
in \cite{jurman08algebraic} to mend this flaw by allowing computing
(Canberra) distance for two lists of different length, still defined
within the framework of the metric methods for permutation groups.
\subsection{Notations}
Let $\mathcal{F}=\{F_j\}_{j=1,\ldots,p}$ be a set of $p$ features, and
let $L$ be a ranked list consisting of $l$ elements extracted (without
replacement) from $\mathcal{F}$.  
If $L=\left(F_{L_1},F_{L_2},\ldots,F_{L_l}\right)$, let $\tau(j)$ be the rank of
$F_j$ in L (with $\tau(F_z)=0$ if $F_z\not\in L$) and define
$\tau=\left( \tau(j) \right)_{j=1,\ldots,p}$ the dual list of $L$.
Consider the set ${S}_L$ of all elements of the symmetric
group $S_\mathcal{F}$ on $\mathcal{F}$ whose top-$l$ sublist is $L$: ${S}_L$
has $(p-l)!$ elements and it is isomorphic to (a coset of) $S_{p-l}$.
Finally, let ${S}_\tau$ be the set of all the dual lists of the
elements in ${S}_L$: if $\alpha\in {S}_\tau$, then $\alpha(i)=\tau(i)$
for all indexes $i\in L$. Thus $S_\tau$ consists of the $(p-|L|)!$
(dual) permutations of $S_p$ coinciding with $\tau$ on the elements
belonging to $L$. Furthermore, note that $\tau(L)=\{1,\ldots,|L|\}$.
\subsection{Shorthands}
If $H_s$ is used to denote the $s$-th armonic number defined
as $\displaystyle{H_s=\sum_{j=1}^s \frac{1}{j}}$, then we can define some useful shorthands: 
\begin{equation*}
\begin{split}
&\Delta(a,b,c) = \sum_{a\leq i\leq b} \frac{|c-i|}{c+i} \\
&=  
\begin{cases}
b-a+1-2c(H_{b+c}-H_{a+c-1}) \\ \mbox{\hspace{12em}} \textrm{if $c<a$}\\
2c(H_{2c}-H_{a+c-1}-H_{b+c}-1)+b+a-1 \\ \mbox{\hspace{12em}} \textrm{if $a\leq c\leq b$}\\
2c(H_{b+c}-H_{a+c-1}) - b+a-1 \\ \mbox{\hspace{12em}} \textrm{if $c>b$}\ , 
\end{cases}
\end{split}
\end{equation*}
\noindent and
\begin{align*}
\varepsilon_k(s) &{} = {} \sum_{j=1}^s jH_{j+k} \\
&{} = {} \frac{(s-k)(s+k+1)}{2} H_{s+k+1} + \frac{k(k+1)}{2} H_{k+1} \\
& + \frac{s(2k-s-1)}{4}
\end{align*}
\begin{align*}
\xi(s) &{} = {} \sum_{j=1}^s (2j) H_{2j}\\
&{} = {} \left(s+\frac{1}{2}\right)^2 H_{2s+1} - \frac{1}{8}H_s -\left(\frac{2s^2+s+1}{4} \right) \ .
\end{align*}
\noindent Finally,
\begin{equation*}
\Theta(\alpha,\beta,\gamma) = \sum_{\alpha\leq u\leq \gamma } \sum_{\beta\leq v\leq \gamma } \frac{|u-v|}{u+v} = \sum_{\alpha\leq u\leq \gamma} \Delta(\beta,\gamma,u)\ ,
\end{equation*}
\noindent with $\Theta(\alpha,\beta,\gamma)=\Theta(\beta,\alpha,\gamma)$.
Details on harmonic number can be found in \cite{graham89concrete}, while some new techniques for dealing with sums and products of harmonic numbers are shown in \cite{cheon07generalized,simic98best,villarino04ramanujan,villarino06sharp,kauers06indefinite,kauers06application,schneider04symbolic,abramov04telescoping,schneider07simplifying}.
\subsection{Canberra distance on permutation groups}
Given two real-valued vectors $\mathbf{x}, \mathbf{y}\in\mathbb{R}^n$, their Canberra distance \cite{lance67mixed} is defined as
\begin{equation*}
\dc(\mathbf{x}, \mathbf{y}) = \sum_{i=1}^{n} \frac{|\mathbf{x}_i- \mathbf{y}_i|}{|\mathbf{x}_i| + |\mathbf{y}_i|}\ .
\end{equation*}
This metric can be naturally extended to a distance on permutation groups: for $\tau,\sigma\in S_p$, we have
\begin{equation*}
\dc(\tau,\sigma) = \sum_{i=1}^{p} \frac{|\tau(i)-\sigma(i)|}{\tau(i)+\sigma(i)}\ .
\end{equation*}
The expected (average) value of the Canberra metric on the whole group $S_p$ can be computed as follows:
\begin{align}
\label{eq:expected}
E\{\dc(S_p)\}&= \frac{1}{\vert S_p\vert^2} \sum_{\sigma,\tau\in S_p} \dc(\sigma,\tau) \nonumber \\ 
&= \frac{1}{\vert S_p\vert} \sum_{\sigma\in S_p} \dc(\sigma,\mathrm{Id}_{S_p}) \nonumber \\
&= \frac{1}{p!} \sum_{\sigma\in S_p} \sum_{i=1}^p \frac{\vert \sigma(i)-i \vert}{\sigma(i)+i} \\
&= \left( 2n+2+\frac{1}{2n}\right) H_{2n}\nonumber \\
&\quad{-}  \left( 2n+2+\frac{1}{4n}\right) H_{n} - \left( n+\frac{3}{2}\right)\nonumber \ .
\end{align}
\subsection{Canberra Distance on Partial Lists}
If $L_1$ and $L_2$ are two (partial) lists of length respectively $l_1\leq l_2$ whose elements belong to $\mathcal{F}$, and $d$ is a distance on permutation groups, define the distance between $L_1$ and $L_2$ as
\begin{align*}
d(L_1,L_2) &= f\left( \left\{ d(\alpha,\beta) \colon \alpha\in{S}_{\tau_1}, \beta\in{S}_{\tau_2} \right\} \right) \\
&= f(d({S}_{\tau_1} ,{S}_{\tau_2} ))\ ,
\end{align*}
for $f$ a function of the $(p-l_1)!(p-l_2)!$ distances
$d(\alpha,\beta)$ such that on a singleton $t$, $f(\{ t \}) = t$.  Note
that if $L_1$ and $L_2$ are complete lists, the above definition
coincides on complete lists with the usual definition of distance
between complete lists given in \cite{jurman08algebraic}.
Moreover, being $d$ a distance, the smaller the value the more similar the compared lists.
A natural choice for the function $f$, motivated also from the fact that the many distances for permutation groups (and we proved this is the case for the Canberra distance in \cite{gobbi08algebraic,jurman09canberra}) are asymptotically normal \cite{hoeffding51combinatorial}, is the mean function, so that
\begin{equation}
\label{eq:definition}
d(L_1,L_2) = \frac{1}{|{S}_{\tau_1}|\cdot |{S}_{\tau_2}|} \sum_{\alpha\in{S}_{\tau_1}} \sum_{\beta\in{S}_{\tau_2}} d(\alpha,\beta)\ .
\end{equation}
We note that this definition differs from the one first introduced in
\cite{jurman08algebraic} because the relation between the size of
actually used features and the size the original feature set is taken
into account here. This is relevant while performing genomic profiling
experiments.
Consider the decomposition of the set $\mathcal{F}$
into the three disjoint sets (ignoring the features' rank)
$F_{12}=L_1{}\cap{}L_2\ $,
$F_{\overline{12}}=\mathcal{F}\setminus\left(L_1{}\cup{}L_2 \right)$
and
$F_{1/2}=\left(L_1{}\cup{}L_2\right)\setminus{}\left(L_1{}\cap{}L_2
\right)$. Then, if $d=\dc$ is the Canberra distance and $\displaystyle{\Lambda= \frac{1}{(p-l_1)!(p-l_2)!}}$, 
the Eq. \eqref{eq:definition} can be split as follows into three terms:
\begin{align}
\label{eq:split}
&\dc(L_1,L_2) = \frac{1}{|S_{\tau_1}|}  \frac{1}{|S_{\tau_2}|} \sum_{\alpha\in{S}_{\tau_1}} \sum_{\beta\in{S}_{\tau_2}} \dc(\alpha,\beta)\nonumber \\
&= \Lambda \sum_{\begin{smallmatrix}\alpha\in S_p, \beta\in S_p \\ \alpha(i)=\tau_1(i)\textrm{ if $i\in L_1$ } \\  \beta(i)=\tau_2(i)\textrm{ if $i\in L_2$ }  \end{smallmatrix}} 
\sum_{i=1}^p \frac{|\alpha(i)-\beta(i)|}{\alpha(i)+\beta(i)}\ ,\nonumber\\
&= \Lambda \sum_{(\alpha, \beta)\in S_{\tau_1}\times S_{\tau_2}} \sum_{F_i\in\mathcal{F}} \frac{|\alpha(i)-\beta(i)|}{\alpha(i)+\beta(i)} \nonumber\\
&=  \Lambda \sum_{F_i\in\mathcal{F}} \sum_{(\alpha, \beta)\in S_{\tau_1}\times S_{\tau_2}} \frac{|\alpha(i)-\beta(i)|}{\alpha(i)+\beta(i)} \nonumber\\
&= \Lambda \sum_{F_i \in F_{12} \cup F_{1/2} \cup F_{\overline{12}}} \sum_{(\alpha, \beta)\in S_{\tau_1}\times S_{\tau_2}} \frac{|\alpha(i)-\beta(i)|}{\alpha(i)+\beta(i)} \nonumber\\
&= \Lambda \left( \sum_{F_i \in F_{12}} \sum_{(\alpha, \beta)\in S_{\tau_1}\times S_{\tau_2}} \frac{|\alpha(i)-\beta(i)|}{\alpha(i)+\beta(i)} \tag{T1} \right.\\
&\qquad{+}\sum_{F_i \in F_{1/2}} \sum_{(\alpha, \beta)\in S_{\tau_1}\times S_{\tau_2}} \frac{|\alpha(i)-\beta(i)|}{\alpha(i)+\beta(i)}\tag{T2} \nonumber\\
&\qquad{+} \left. \sum_{F_i \in F_{\overline{12}}} \sum_{(\alpha, \beta)\in S_{\tau_1}\times S_{\tau_2}} \frac{|\alpha(i)-\beta(i)|}{\alpha(i)+\beta(i)}\right) \tag{T3}\nonumber\ .
\end{align}
Expanding the three terms T1, T2, T3 a closed form for the distance can be reached:
\begin{definition}[Complete Canberra Distance.]
\label{def:complete}
The Complete Canberra Distance (between partial lists) is defined as
\begin{align}
\label{eq:final}
\dc(L_1,L_2)&= \sum_{i\in L_1 \cap L_2} \left( \frac{| \tau_1(i)-\tau_2(i)|}{\tau_1(i)+\tau_2(i)} \right. \nonumber\\
&\quad\qquad\left.{-}\: \frac{\Delta(l_2+1,p,\tau_1(i))}{p-l_2} \right. \\
&\quad\qquad\left.{-}\:  \frac{\Delta(l_1+1,p,\tau_2(i))}{p-l_1} \right) \nonumber\\
&\qquad{+}\: \frac{1}{p-l_2} \left( l_1(p-l_2) -2\varepsilon_p(l_1) +2\varepsilon_{l_2}(l_1)\right) \nonumber\\
&\qquad{+}\: \frac{1}{p-l_1} \left( l_1(p-l_1) +4\varepsilon_{l_1}(l_1) + 2\xi(l_2)  \right.\nonumber\\
&\qquad\left. {-}\: 2\xi(l_1)- 2\varepsilon_{l_1}(l_2) -2\varepsilon_p(l_2)\right.\nonumber\\
&\qquad\left. {+}\: (p+l_1)(l_2-l_1) + l_1(l_1+1)\right.\nonumber\\
&\qquad\left. {-}\: l_2(l_2+1) \right) \nonumber\\
&\qquad{+}\: A\cdot \left( 2\xi(p)-2\xi(l_2)-2\varepsilon_{l_1}(p)+ 2\varepsilon_{l_1}(l_2)\right. \nonumber\\
&\qquad\left.{-}\: 2\varepsilon_p(p) +2\varepsilon_p(l_2) +(p+l_1)(p-l_2)\right.\nonumber\\
&\qquad\left.{+}\: l_2(l_2+1)-p(p+1) \right)\ , \nonumber
\end{align}
\noindent where $\displaystyle{A = \frac{|\mathcal{F}\setminus (L_1\cup L_2) |}{(p-l_1)(p-l_2)}}$ .
\end{definition}
The sum generating the term T3 in Eq. \eqref{eq:split} runs over the subset $F_{\overline{12}}$ collecting all elements of the original feature set which do not occur neither in the first list nor in the second. 
Thus this part of the formula is independent from the positions of the elements occurring in the partial lists $L_1$, $L_2$. 
Neglecting this term, we obtain another measure of list difference:
\begin{definition}[Core Canberra Distance.]
The Core Canberra is defined as the components T1, T2 of the Complete Canberra Distance depending on the positions of the elements in the considered partial lists. 
This corresponds to setting $A=0$ in Eq. \eqref{eq:final} of Def. \ref{def:complete}.
\end{definition}

Throughout the paper, the values of both instances of the Canberra Distance are normalized by dividing them by the expected value $E\{\dc(S_p)\}$ on the whole permutation group $S_p$ reported in Eq. \eqref{eq:expected}. This would result in two random (complete) lists having a Complete Canberra Distance very close to one; note that, since the expected value is not the highest one, distance values greater than one can occur.
When the number of features in $\mathcal{F}$ not occurring in $L_1$, $L_2$ becomes larger, the non-core component gets numerically highly preeminent: in fact, in the term T3 all the possible $(p-l1)!(p-l2)!$ lists in $S_p$ having $L_1$ and $L_2$ respectively as top lists are considered; as an example, for $p=10000$ and $L_1$, $L_2$ two partial lists with $100$ elements, this corresponds in evaluating the distance among $9900!^2\simeq 2.2\cdot 10^{70519}$ lists of elements not occurring in $L_1$, $L_2$. When the number of lists of unselected elements grows larger, the average distance among them would get closer to the expected value of the Canberra distance on $S_p$ because of the Hoeffding's Theorem. 

This is quite often the case for biological ranked lists: for instance, selecting a panel of biomarkers from a set of probes usually means choosing less than hundred of features out of an original set of several thousands.  
Thus, considering the Core component instead of the Complete Distance can be helpful in term of dimensionality reduction of the considered problem. 

\begin{figure}[t]
\begin{center}
\textbf{a}
\includegraphics[width=0.2\textwidth]{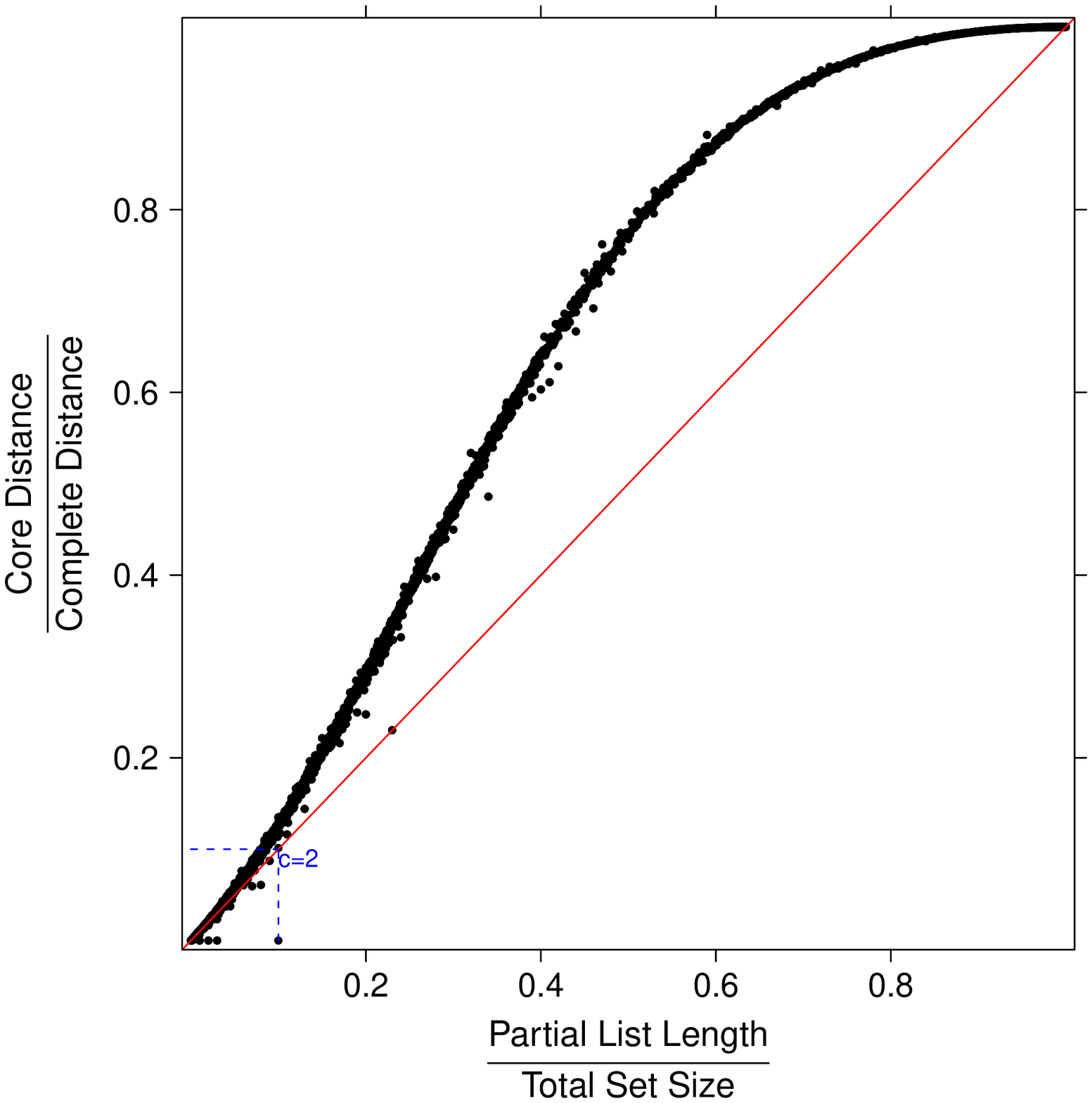}
\includegraphics[width=0.2\textwidth]{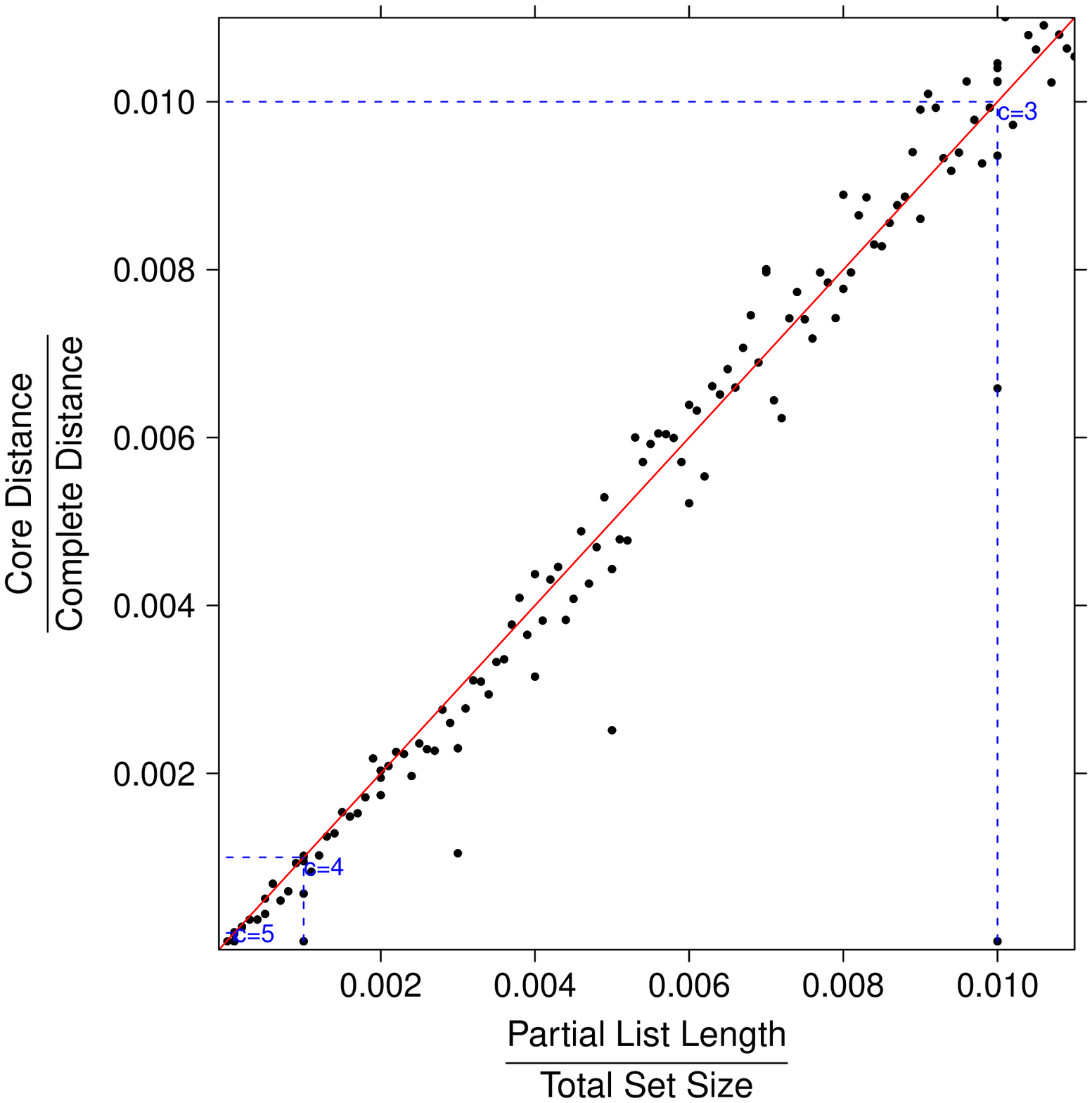}
\textbf{b}
\end{center}
\caption{\textbf{(a)} Ratio between Core and Complete distances versus ratio between the length of partial lists and the size of the full feature set for about 7000 instances of couples of partial lists. Lists pairs have the same length and they are randomly permuted, with partial lists length ranging between 1 and 5000 and full set size ranging between 10 and 100000. \textbf{(b)} Zoom of the bottom left corner of panel \textbf{(a)}: Core and Complete distances are proportional when the ratio between the length of partial lists and the size of the full feature set is less than about 0.15.} 
\label{fig:core_comp}
\end{figure}

As an example, in Tab. \ref{tab:core_comp} we show the values of the normalized distances of two partial lists of length $10$ extracted from an original set $\mathcal{F}$ with $p=10^c$ features ($c=2,3,4,5$), in the three cases of identical partial lists, randomly permuted partial lists (which yields average distance) and maximally distant partial lists (see \cite{jurman09canberra, gobbi08algebraic} for the identification of the permutation maximizing the Canberra distance between lists).
\begin{table}[tb]
\tiny
\begin{center}
\begin{tabular*}{0.73\columnwidth}{llrrrr}
\hline
Lists &Dist.  & \multicolumn{1}{c}{$c=2$} & \multicolumn{1}{c}{$c=3$} & \multicolumn{1}{c}{$c=4$} & \multicolumn{1}{c}{$c=5$} \\ 
\hline
Identical & Comp. & 0.692830 & 0.960499 & 0.995858 & 0.999583 \\
Random    & Core     & 0.078038	& 0.006368 & 0.000950 & 0.000109 \\
Random 	  & Comp. & 0.770868 & 0.966867 & 0.996809 & 0.999692 \\
Max.Dist. & Core     & 0.128448 & 0.012665 & 0.001265 & 0.000126 \\
Max.Dist. & Comp. & 0.821278 & 0.973164 & 0.997123 & 0.999709 \\
\hline
\end{tabular*}
\end{center}
\caption{Core and Complete normalized Canberra distance for two partial lists of 10 features extracted from a set of $|\mathcal{F}|={10}^c$ features. The partial lists are either identical, randomly permuted (average distance) or maximally distant. The Core Distance for Identical partial lists is zero.}
\label{tab:core_comp}%
\normalsize
\end{table}
A further observation can be derived from Fig. \ref{fig:core_comp}, where the ratio between Core and Canberra distances are plotted versus the ratio between the length of partial lists and the size of the full feature set for about 7000 instances of couples of partial lists of the same length randomly permuted.
When the number of elements of the partial lists is a small portion of the total feature, the Complete and the Core distance are almost linearly dependent, while when such ratio approaches one the ratio between the two measures follows a different function.
As shown above, the Core measure is more convenient to better focus on differences occurring among lists of relatively small length. On the other hand, the Complete version is the elective choice when the original feature set is large and the partial lists' length is of comparable order of magnitude of $|\mathcal{F}|$.
\subsection{Expansion of Eq. \eqref{eq:split}}
The three terms occurring in Eq. \eqref{eq:split} can be expanded through a few algebraic steps in a more closed form, reducing the use of sums wherever possible.
\subsubsection{T1: common features}
The first term is the component of the distance computed over the features appearing in both lists $L_1$, $L_2$, thus no complete closed form can be written. The expansion reads as follows:
\begin{align*}
&\sum_{F_i \in F_{12}} \sum_{(\alpha, \beta)\in S_{\tau_1}\times S_{\tau_2}}  \frac{|\alpha(i)-\beta(i)|}{\alpha(i)+\beta(i)}=\\
&=\sum_{i\in L_1\cap L_2}  \sum_{(\alpha, \beta)\in S_{\tau_1}\times S_{\tau_2}} \frac{|\alpha(i)-\beta(i)|}{\alpha(i)+\beta(i)} \\
&= \sum_{i\in L_1\cap L_2} |S_{\tau_1}|\cdot |S_{\tau_2}| \frac{|\tau_1(i)-\tau_2(i)|}{\tau_1(i)+\tau_2(i)}\\
&= {(p-l_1)!(p-l_2)!}  \sum_{i\in L_1\cap L_2}   \frac{|\tau_1(i)-\tau_2(i)|}{\tau_1(i)+\tau_2(i)}\\
&= \Lambda^{-1}  \sum_{i\in L_1\cap L_2}   \frac{|\tau_1(i)-\tau_2(i)|}{\tau_1(i)+\tau_2(i)}\ .
\end{align*}
\subsubsection{T2: features occurring only in one list}
The second term regards the elements occurring only in one of the two lists. By defining $\lambda_1=(p-l_1)!(p-l_2-1)!$ and $\lambda_2=(p-l_2)!(p-l_1-1)!$, the term can be rearranged as:
\begin{align*}
&\sum_{F_i \in F_{1/2}} \sum_{(\alpha, \beta)\in S_{\tau_1}\times S_{\tau_2}} \frac{|\alpha(i)-\beta(i)|}{\alpha(i)+\beta(i)} = \\
&= \sum_{i\in L_1, i\not\in L_2} \sum_{(\alpha, \beta)\in S_{\tau_1}\times S_{\tau_2}} \frac{|\alpha(i)-\beta(i)|}{\alpha(i)+\beta(i)} \\
&\qquad{+} \sum_{i\in L_2, i\not\in L_1} \sum_{(\alpha, \beta)\in S_{\tau_1}\times S_{\tau_2}} \frac{|\alpha(i)-\beta(i)|}{\alpha(i)+\beta(i)} \\
&=\sum_{i\in L_1, i\not\in L_2} \sum_{\beta\in S_{\tau_2}} |S_{\tau_1}| \frac{|\tau_1(i)-\beta(i)|}{\tau_1(i)+\beta(i)}\\ 
&\qquad{+} \sum_{i\in L_2, i\not\in L_1} \sum_{\alpha\in S_{\tau_1}} |S_{\tau_2}| \frac{|\alpha(i)-\tau_2(i)|}{\alpha(i)+\tau_2(i)} \\
&=\lambda_1 \sum_{i\in L_1, i\not\in L_2} \sum_{j= l_2+1}^p \frac{|\tau_1(i)-j|}{\tau_1(i)+j} \\
&\quad{+} \lambda_2 \sum_{i\in L_2, i\not\in L_1} \sum_{j= l_1+1}^p \frac{|j-\tau_2(i)|}{j+\tau_2(i)}\\
&=\lambda_1 \sum_{i\in L_1, i\not\in L_2} \Delta(l_2+1,p,\tau_1(i)) \\
&\quad{+}\lambda_2 \sum_{i\in L_2, i\not\in L_1} \Delta(l_1+1,p,\tau_2(i))\\
&=\lambda_1 \left( \sum_{i\in L_1} \Delta(l_2+1,p,\tau_1(i)) \right.\\
&\qquad{-} \left. \sum_{F_i\in F_{12}} \Delta(l_2+1,p,\tau_1(i))  \right)\\
&\quad{+} \:  \lambda_2 \left( \sum_{i\in L_2} \Delta(l_1+1,p,\tau_2(i)) \right. \\
&\qquad{-} \left. \sum_{F_i\in F_{12}} \Delta(l_1+1,p,\tau_2(i)) \right) \ .
\end{align*}
\subsubsection{T3: unselected features}
The last term represents the component of the distance computed on the factor group, that is the elements of the original feature set not appearing in $L_1$ or $L_2$. 
Here a complete closed form can be reached:
\begin{align*}
&\sum_{F_i \in F_{\overline{12}}}  \sum_{(\alpha, \beta)\in S_{\tau_1}\times S_{\tau_2}} \frac{|\alpha(i)-\beta(i)|}{\alpha(i)+\beta(i)} = \\
&=\sum_{i\not\in L_1\cup L_2} \sum_{(\alpha, \beta)\in S_{\tau_1}\times S_{\tau_2}} \frac{|\alpha(i)-\beta(i)|}{\alpha(i)+\beta(i)} \\
&= |F_{\overline{12}}| (p-l_1-1)! (p-l_2-1)! \sum_{i=l_1+1}^p \sum_{j=l_2+1}^p \frac{|i-j|}{i+j}\\
&= |F_{\overline{12}}| (p-l_1-1)! (p-l_2-1)!\Theta(l_1+1,l_2+1,p)\\ 
&= \frac{|\mathcal{F}\setminus (L_1\cup L_2) |}{(p-l_1)(p-l_2)}\cdot\left( 2\xi(p)-2\xi(l_2)-2\varepsilon_{l_1}(p)\right. \\
&\qquad\left.{+}\: 2\varepsilon_{l_1}(l_2) - 2\varepsilon_p(p) +2\varepsilon_p(l_2) +(p+l_1)(p-l_2)\right.\\
&\qquad\left.{+}\: l_2(l_2+1)-p(p+1) \right)\ .
\end{align*}
\subsection{The Borda list}
To summarize the information coming from a set of lists $\mathcal{L}$ into a single optimal list we adopt the same strategy of \cite{jurman08algebraic}, \textit{i.e.} an extension of the classical voting theory methods known as the Borda count \cite{borda81memoire,saari01chaotic}. 
This method, in its basic version, derives a single list from a set of $B$ lists on $p$ candidates $F_1,\ldots, F_p$ by ranking them according to a score $s(F_i)$ defined by the total number of candidates ranked higher than $F_i$ over all lists.
Our extension consists in first computing, for each feature $F_j$, its number of extractions (the number of lists where $F_j$ appears) $\displaystyle{e_j=|\{i \in \{1\ldots B\} \colon F_j\in L_i\}|}$ and its average position number $\displaystyle{a_k(j)=\frac{1}{e_j}\sum_{\{i \in \{1\ldots B\} \colon F_j\in L_i\}} \tau_i(j)}$ and then ranking the features by decreasing extraction number and by increasing average position number when ties occur. The resulting list will be called optimal list or Borda list.
In \cite{jurman08algebraic} the equivalence of this ranking with the Borda count is proved.
\subsection{Implementation}
The computation of the stability indicator for partial lists is publicly available (since version
1.1.2) within the Open Source Python package \texttt{mlpy}
(\url{https://mlpy.fbk.eu}) for statistical machine learning \cite{albanese08nips,albanese09mlpy}.
\subsection{Data description}
For the experiments described in the RESULTS, we used two datasets: a synthtetic dataset and a microarray dataset.
\subsubsection{The Tib datasets.}
We build two datasets simulating a microarray
dataset inspired by \cite{bair04semi}. The datasets Tib100
and Tib500 consist of 100 samples and are described
respectively by 100 and 500 features (genes). The first 50 samples
were assigned to class 1, the others to class -1. All expression
values were generated as standard normally distributed numbers. Genes
1-20 (1-50) have mean 1 for samples 1-50 and mean -1 for samples
51-100.  Initially, genes 21-100 (51-500) in all the samples have mean
0.  Then three substitutions are performed, where a percentage $P$ of
all genes from the $a$-th to the $b$-th are replaced by normally
distributed numbers with mean $m$, namely:
\begin{enumerate}
\item $P=40$, $a=21$, $b=40$ ($50$) and $m=-1$;
\item $P=50$, $a=41$ ($51$), $b=60$ ($150$) and $m=1$;
\item $P=70$, $a=61$ ($151$), $b=70$ ($250$) and $m=0.5$.
\end{enumerate}
While only the first 20 genes are truly discriminating, the noisy part of the dataset is modified in order to give a partial discriminating power also to genes 21-70 (21-250), leaving only the genes 71-100 (251-500) as undiscriminative features.

\subsubsection{The Prostate Cancer dataset.}
We use the publicly available prostate cancer dataset described in \cite{setlur08estrogen}
and available from GEO (accession number GSE8402) built from a custom
Illumina DASL Assay of 6144 genes known from literature to be prostate
cancer related. Setlur et al.  identified a subtype of prostate cancer
characterized by the fusion of the 5'-untranslated region of the
androgen-regulated transmembrane protease serine 2 (TMPRSS2) promoter
with erythroblast transformation-specific transcription factor family
members (TMPRSS2-ER). As mentioned in the original paper, ''the common
fusion between TMPRESS2 and v-ets erythroblastosis virus E26 oncogene
homolog (avian) (ERG) is associated with a more aggressive clinical
phenotype, implying the existence of a distinct subclass of prostate
cancer defined by this fusion''. The discrimination task consists in
separating positive TMPRSS2-ERG gene fusion cases from negative ones.
The database includes two different cohorts of patients: the US
Physician Health Study Prostatectomy Confirmation Cohort, with 41 positive and 60 negative samples, and the
Swedish Watchful Waiting Cohort, consisting of 62 positive and 292 negative samples. 
In what follows, we will indicate the whole dataset as Setlur, and its two cohorts by the shorthands US and Sweden.

\section{RESULTS}
\label{sec:results}%
Two applications are shown in the present section as practical examples of use of the proposed method within common tasks in computational biology. 
First we outline how to use the Canberra distance to compare the different behaviours of several filtering methods on a synthetic dataset. 
Identifying the genes which are differentially
expressed between two groups of samples is a key task in a profiling
study: when the sample size is small this may be quite tricky, since
the chances of selecting false positives are relevant. Many algorithms
have been devised to deal with such issue: an important family is
represented by the filter methods, which essentially consist in
applying a suitable statistic to the dataset to rank the genes in term
of a degree of differential expression, and then deciding a threshold
(cutoff) on such degree to discriminate the differentially expressed
genes.
Reliability of a method over another is a debated issue in literature:
while some authors thinks that the lists coming from using FC ratio
are more reproducible than those emerging by ranking genes according
to the $P$-value of $t$-test
\cite{yao08disease,chen07reproducibility}, others
\cite{simon08microarray} point out that $t$-test and $F$-test better
address some FC deficiencies (e.g. ignoring variation within the same
class) and they are recommended for small sample size datasets.  Most
researcher also agree on the fact that SAM
\cite{tusher01significance,storey02direct,efron01empirical,efron02empirical,taylor05miss}
should outperform all other three methods because of its ability in
controlling the false discovery rate. Moreover, in
\cite{witten07comparison} the author show that motivation for the use
of either FC or mod-$t$ is essentially biological while ordinary t
statistic is shown to be inferior to the mod-$t$ statistic and
therefore should be avoided for microarray analysis.
In the extensive study \cite{jeffery06comparison}, also alternative methods such as 
Empirical Bayes Statistics, Between Group Analysis and Rank Product have been taken into account, 
applying them to 9 microarray publicly available datasets. The
resulting gene lists are compared only in terms of number of
overlapping genes and predictive performance when using as features to
train four different classifier.
Here we will study the stability of the lists of discriminative genes produced by several filtering algorithms as a function of the number of samples, by evaluating it at different values of the filtering thresholds.
Our second application is a profiling task on a publicly available recent prostate cancer dataset, where we aim at detecting a panel of genes involved in the discrimination between patients expressing or not a certain gene fusion. 
We use the ranked partial lists produced by replicated cross-validations to better characterize the seeked panel and to detect differences between the two cohorts in the dataset. 
Finally, we compare the set of ranked lists produced by the profiling experiment with the sets of lists retrieved by applying the same seven filtering methods to the prostate cancer dataset, to show similarities and differences of lists obtained when looking for a discriminative predictive panel and when identifying differentially expressed genes.
\subsection{Gene Filtering on the Tib datasets}
\label{ssec:filtering}
\begin{figure*}[bth!]
\begin{center}
\textbf{(a)}
\includegraphics[width=0.45\textwidth]{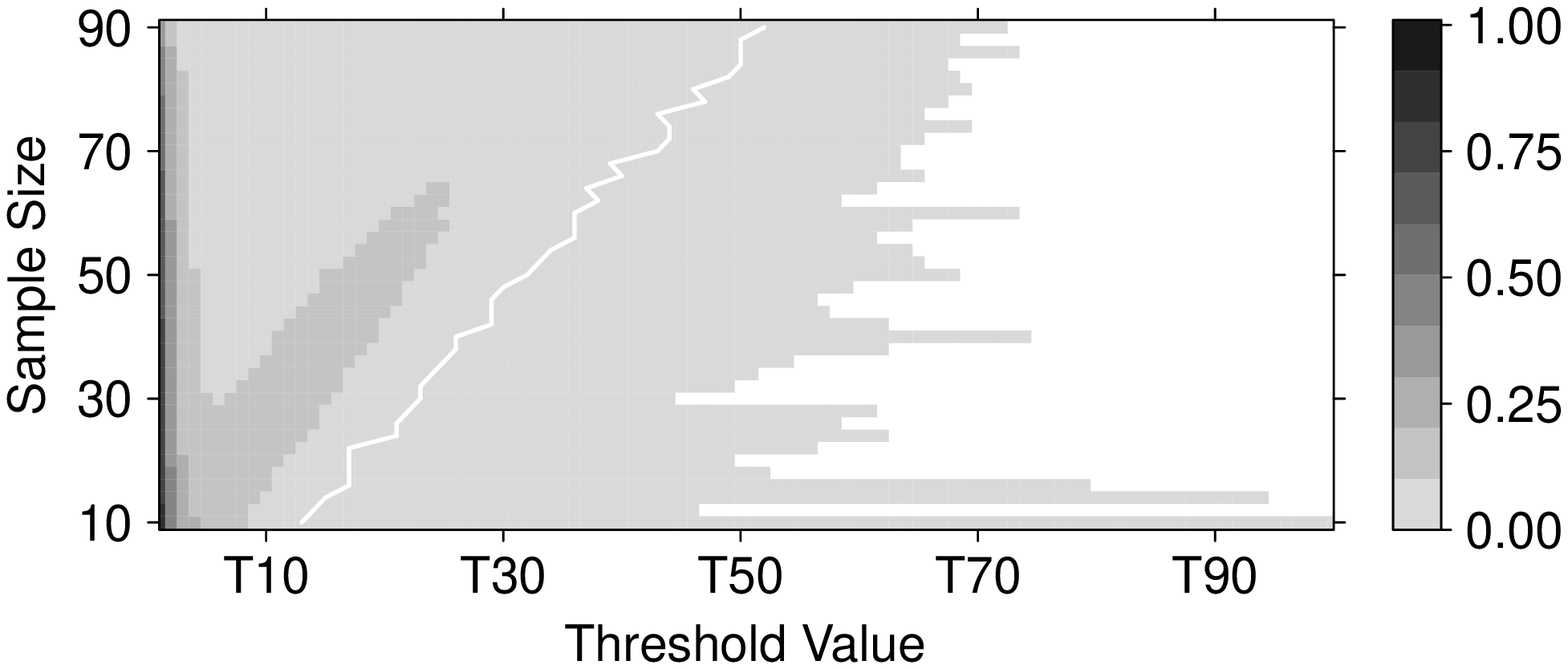}
\includegraphics[width=0.45\textwidth]{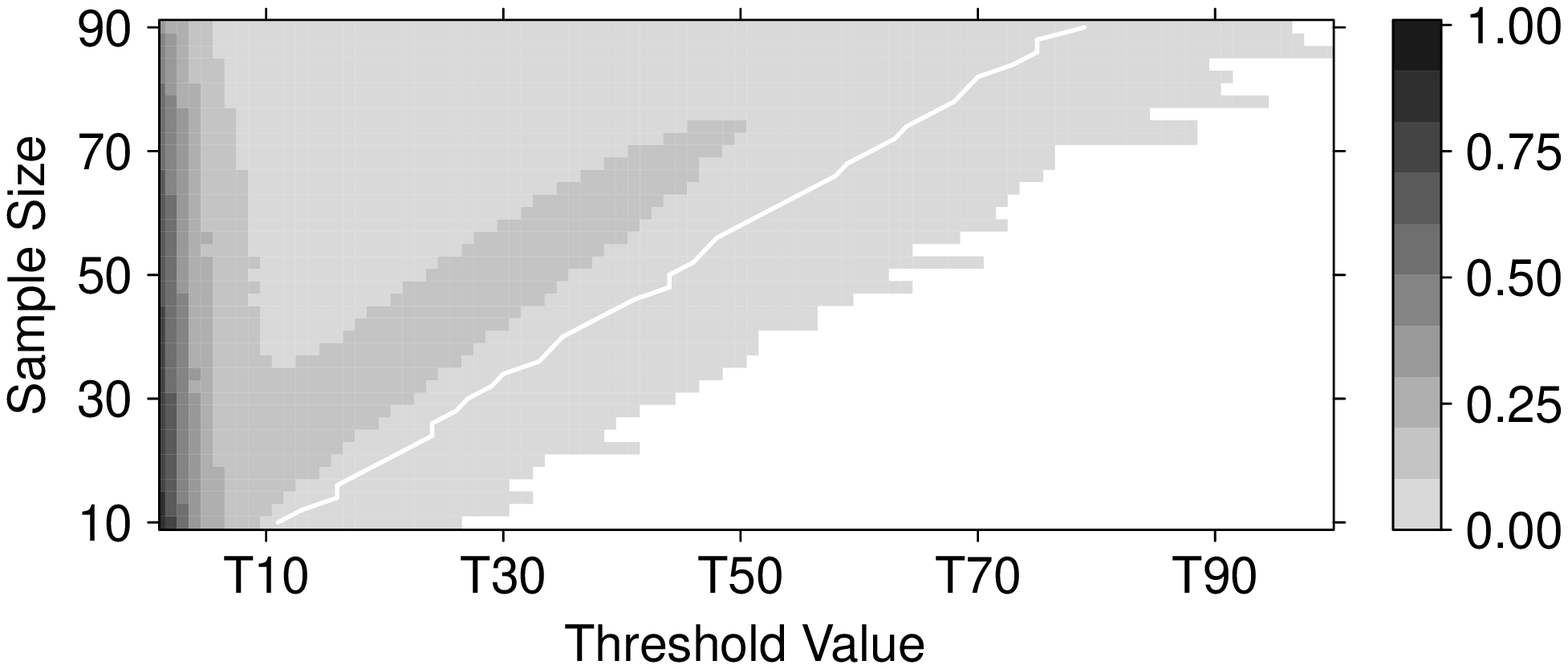}
\textbf{(b)}
\\
\textbf{(c)}
\includegraphics[width=0.45\textwidth]{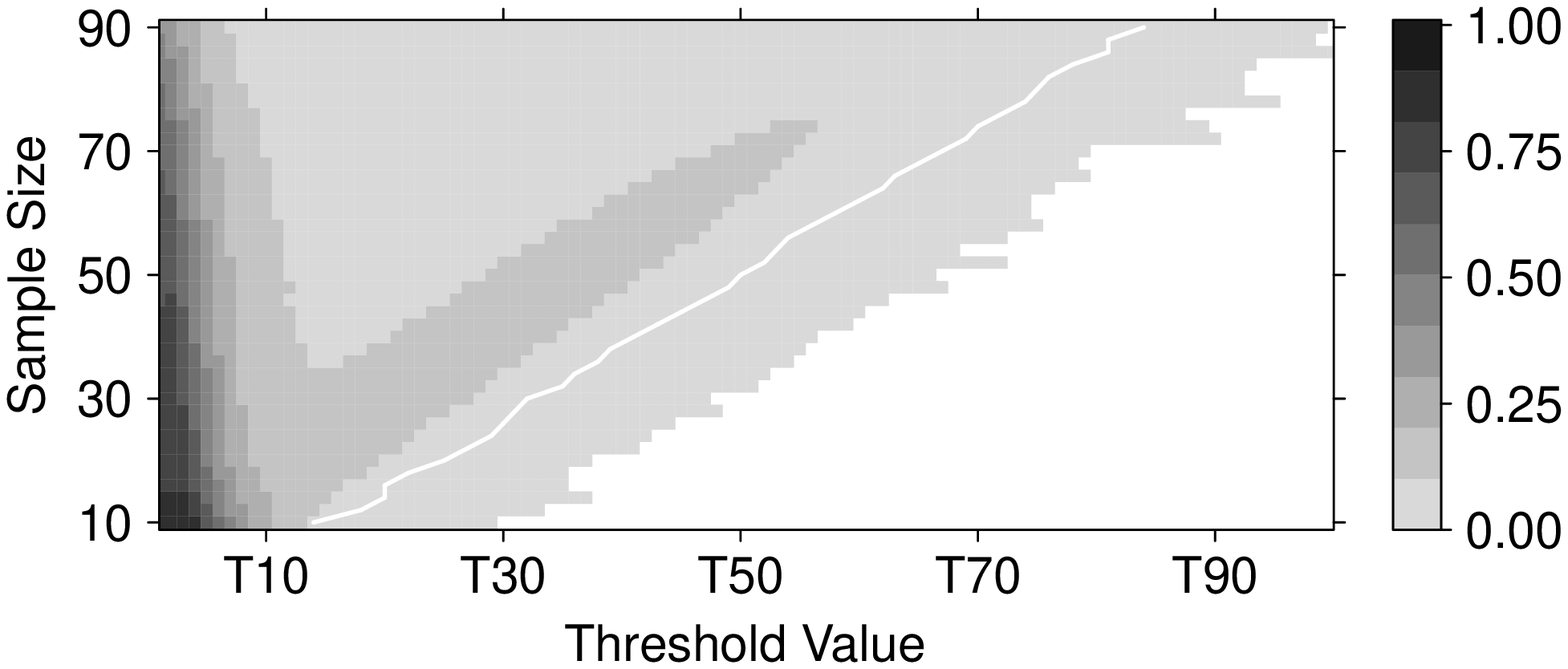} 
\includegraphics[width=0.45\textwidth]{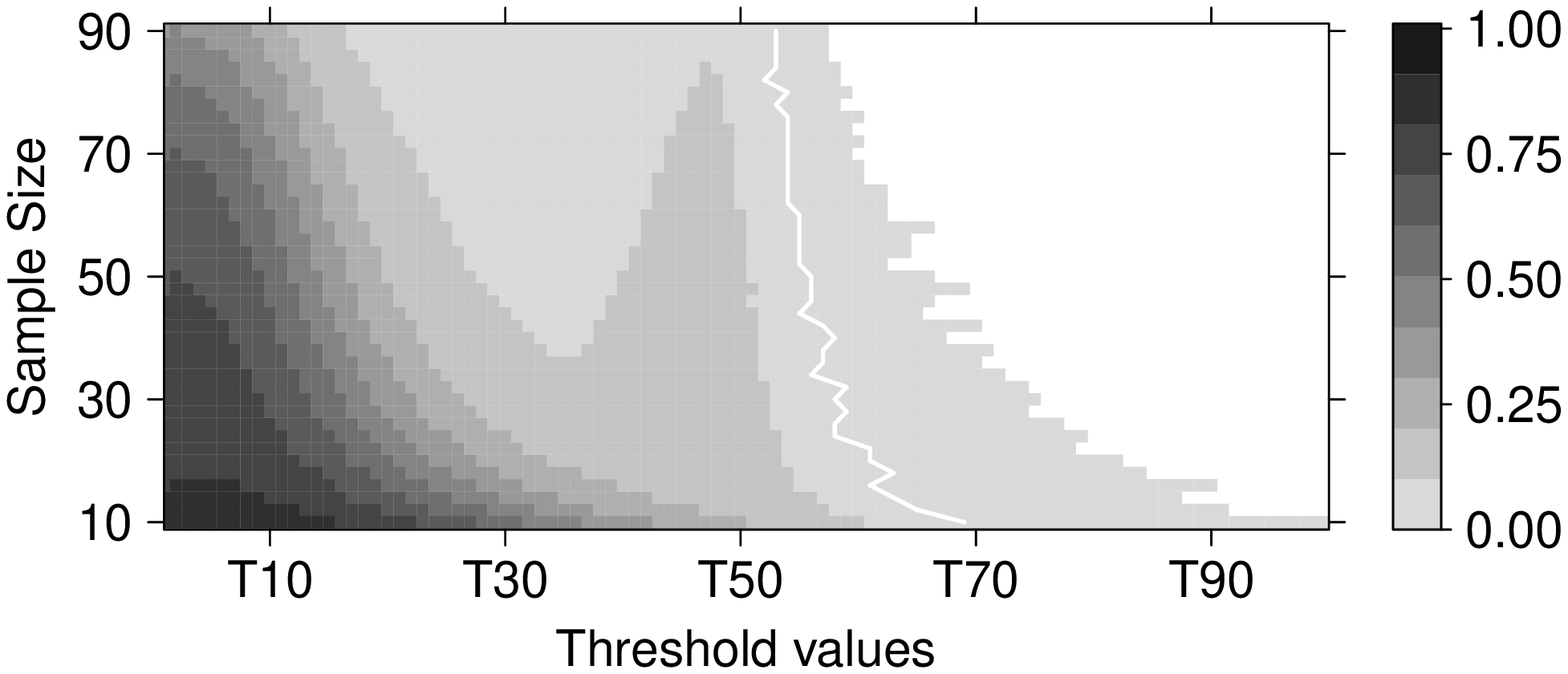} 
\textbf{(d)}
\\
\textbf{(e)}
\includegraphics[width=0.45\textwidth]{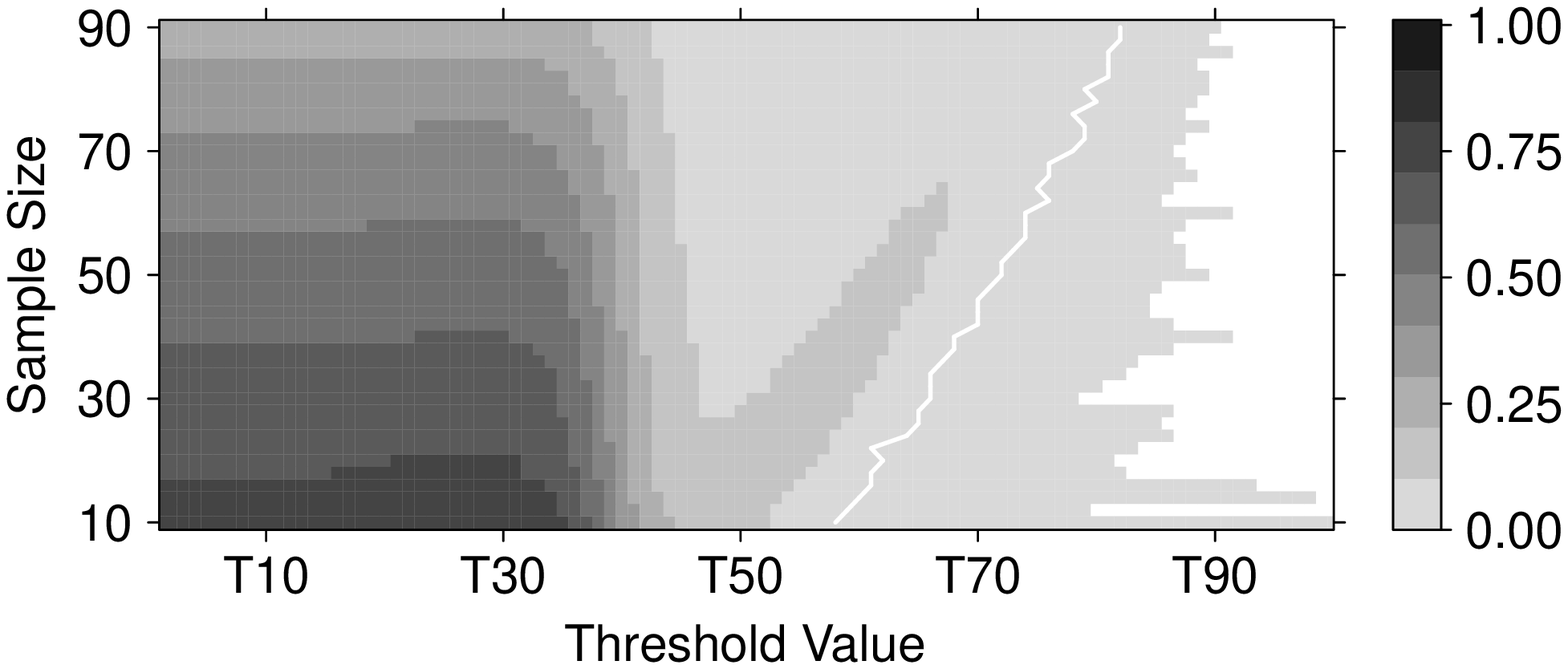}
\includegraphics[width=0.45\textwidth]{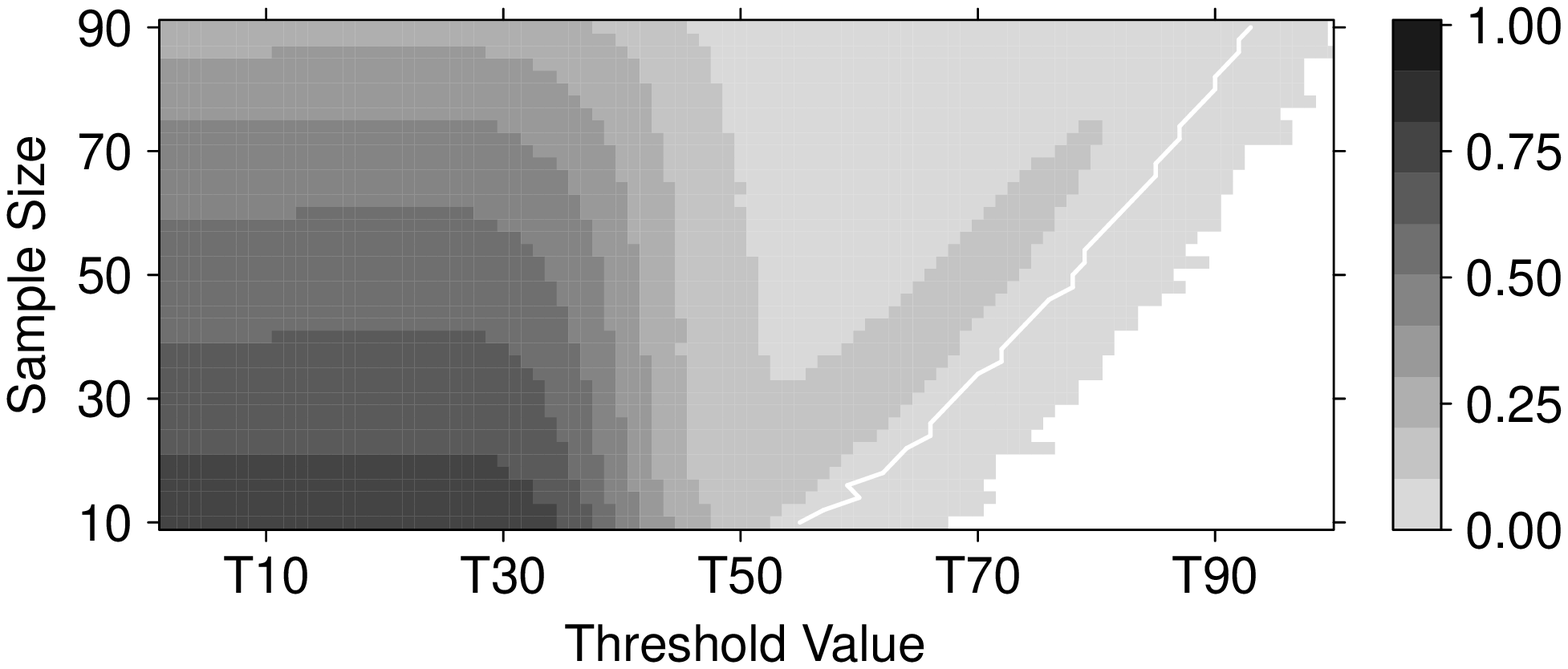} 
\textbf{(f)}
\\
\textbf{(g)}
\includegraphics[width=0.45\textwidth]{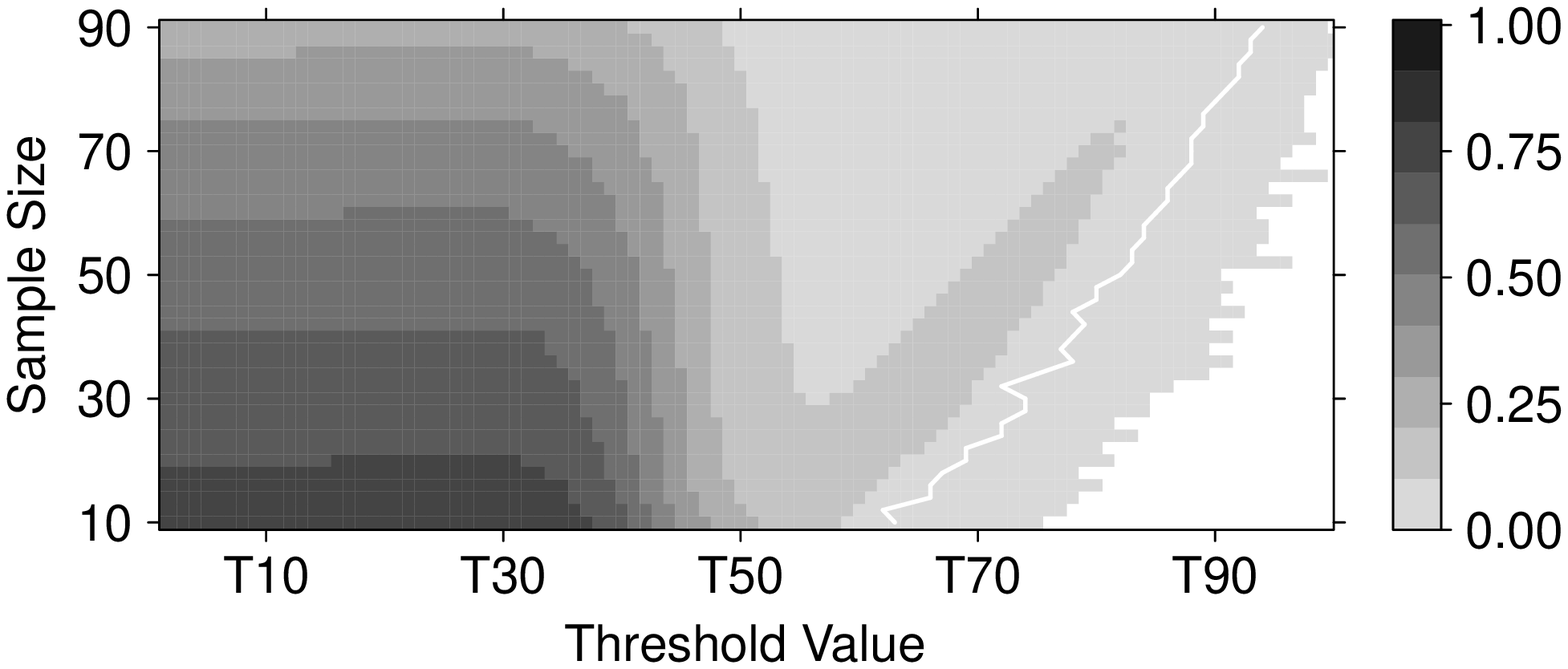}
\includegraphics[width=0.45\textwidth]{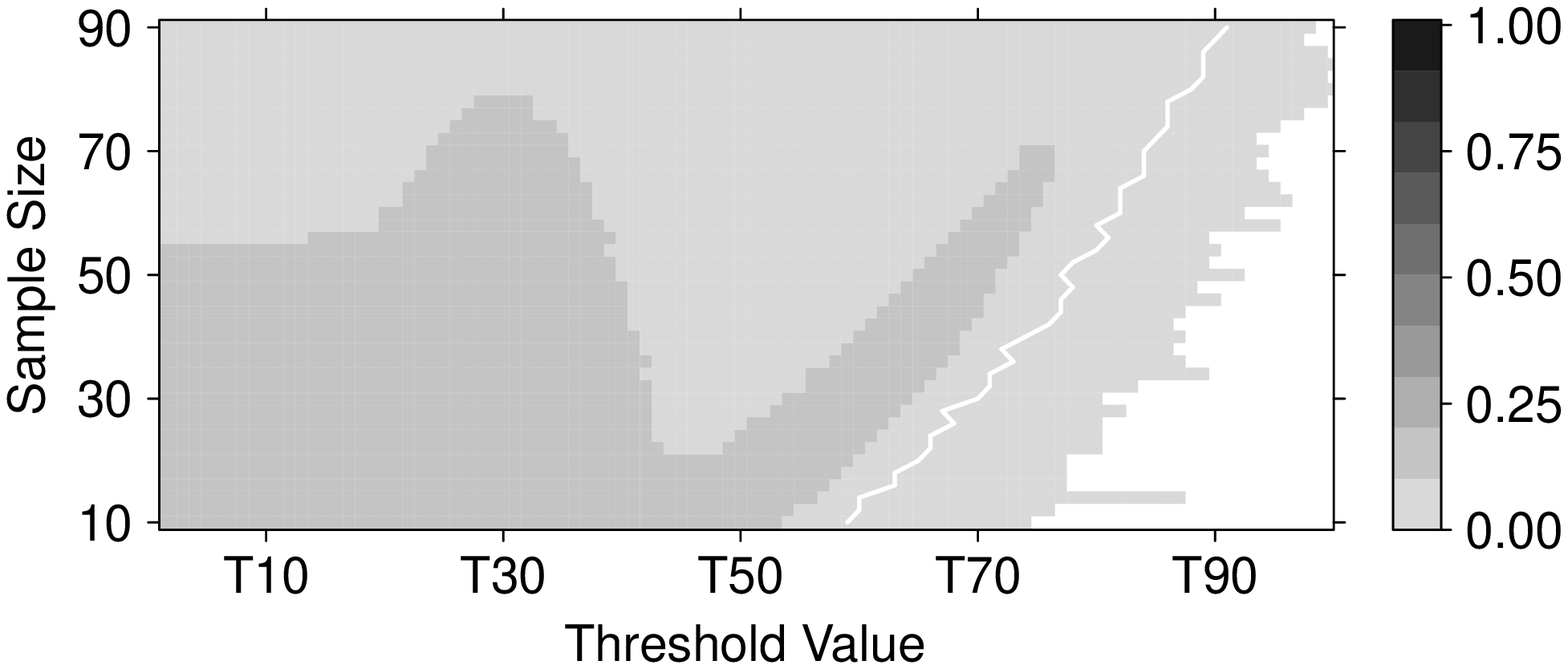}
\textbf{(h)}
\\
\end{center}
\caption{
Levelplot of the Core Canberra distance evaluated on the list sets
$L(n,\mathcal{A},\theta,i)$.  White pixels indicates experiments in which all
features are discarded from all lists ($i=0$). The white line
separates list sets with $i=B$ (left side) from those with $i<B$
 (right side). \textbf{(a)}-\textbf{(g)}: Tib500, \textbf{(h)}: Tib100; 
\textbf{(a)}: $F$, \textbf{(b)}: mod-$F$, \textbf{(c)}: B, \textbf{(d)}: FC, \textbf{(e)}: $t$, \textbf{(f)}: mod-$t$, \textbf{(g)}-\textbf{(h)}: SAM.
}
\label{fig:filtering-synt}
\end{figure*}

The stability (i.e.  robustness against input variation) of the gene lists produced by
different filtering strategies on the two Tib100 and Tib500 datasets and computed for different configurations of two parameters (the number of samples and the filtering threshold) is assessed through the experiment outlined in the present section.
By using the stability indicator defined, we explore the properties of
7 state-of-the-art filtering approaches in terms of homogeneity of the
ordered lists of features identified as differentially expressed on
two synthetic datasets. The statistics considered are Fold Change (FC)
\cite{tusher01significance}, Significance Analysis of Microarray (SAM)
\cite{tusher01significance}, $B$ statistics
\cite{loennsted01replicated}, $F$ statistics \cite{neter96applied}, $t$
statistics \cite{jeffery06comparison}, and mod-$F$ and mod-$t$
statistics \cite{smyth04linear}, which are the moderated version of
$F$ and $t$ statistics. The FC of a given gene
is defined here as the ratio of the average expression value
computed over the two groups of samples.
We apply the stability indicators to list sets
$L(n,\mathcal{A},\theta)$ of cardinality $B=100$, where
\begin{itemize}
\item $n$ is the number of samples of each class selected from the original dataset considered in the stability analysis (for each experiment we consider a subset of the original dataset of cardinality equal to $2n$): $n$ ranges between 5 and 45;
\item $\mathcal{A}$ indicates one of the 7 filtering statistics: FC, SAM, $B$ statistics, F statistics, $t$ statistics, and mod-$F$ and mod-$t$ statistics;
\item $\theta$ is the threshold considered for $\mathcal{A}$ so that a set of 100 values was chosen for each $\mathcal{A}$ as a percentage of the $\mathcal{A}$ range.
\end{itemize}
We indicate also as $i=i(\theta)$ the number of elements of the list
set ($i{}<{}B$) and for representation's clearness we will consider
$L(n,\mathcal{A},\theta,i)$. For each parameter configurations we compute the Core Canberra distance.
All filtering statistics are computed by using the package DEDS
\cite{xiao08bioconductor} for BioConductor
\cite{gentleman04bioconductor} within the statistical environment R
\cite{R2008}.
In Fig.~\ref{fig:filtering-synt} we represent the value of the Core 
Canberra distance for some of the values of the triplet (dataset,
$\mathcal{A}$, measure). A few consideration can be drawn by observing the reported images.
First of all, three groups of different behaviours can be identified: $F$, mod-$F$ and $B$ group together and $t$, mod-$t$ and $SAM$ do the same, while $FC$ exhibits a completely different shape. 
In both cases, the mod statistic (both $F$ and $t$) belongs to the same group and it has a better regularization than the corresponding classic counterpart, resulting more robust in the small sample size case.

The group including $t$, mod-$t$, and SAM has a small number of void lists even for high thresholds, and the slope of its white line is higher: both facts indicate a not so strong dependence on the number of samples considered. On the other hand, when the same threshold is considered, the stability is higher for the group $F$, mod-$F$, and $B$. 
The levelplot \ref{fig:filtering-synt}d for $FC$ shows that the dependence of this methods on the dataset sample size is opposite to the behaviour of all other methods: at a given threshold, the constraint imposed gets stricter for increasing number of samples.
The darker horn-shaped area in the rightmost zone of the plots is probably due to the effect that the relevant features come in groups because of the dataset definition, and this is mostly evident in the small sample sizes.
As a final consideration, plots \ref{fig:filtering-synt}g and \ref{fig:filtering-synt}h show that considering the smaller dataset Tib100 instead of the larger Tib500 reflects in losing some details in the corresponding levelplot.

A few more computations and considerations on the stability of the obtained lists are shown and discussed in the following subsections.

\subsection{Profiling}
\label{ssec:profiling}
\begin{figure*}[tb]
\begin{center}
\textrm{a}
\includegraphics[width=0.45\textwidth]{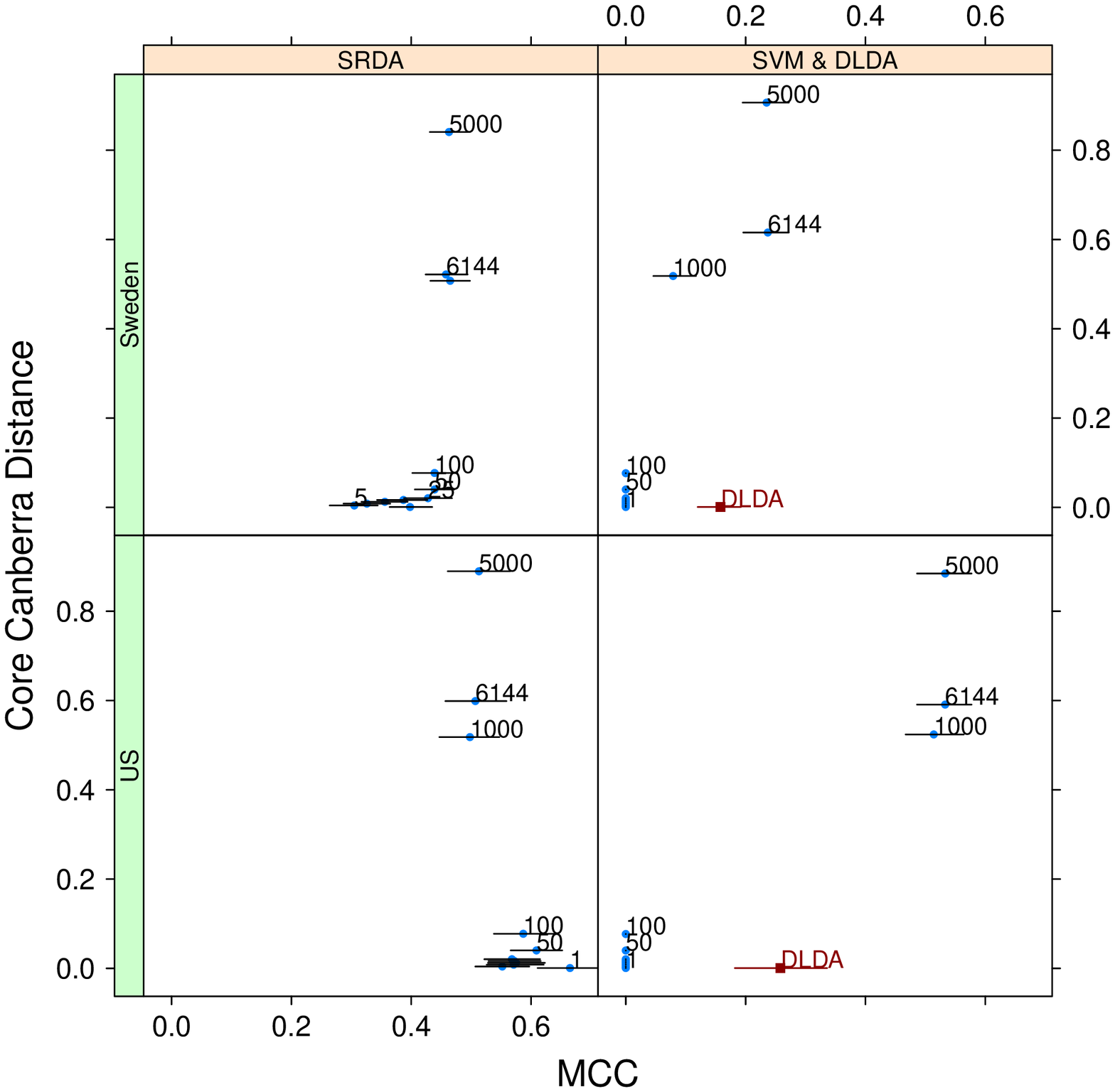}
\includegraphics[width=0.28\textwidth]{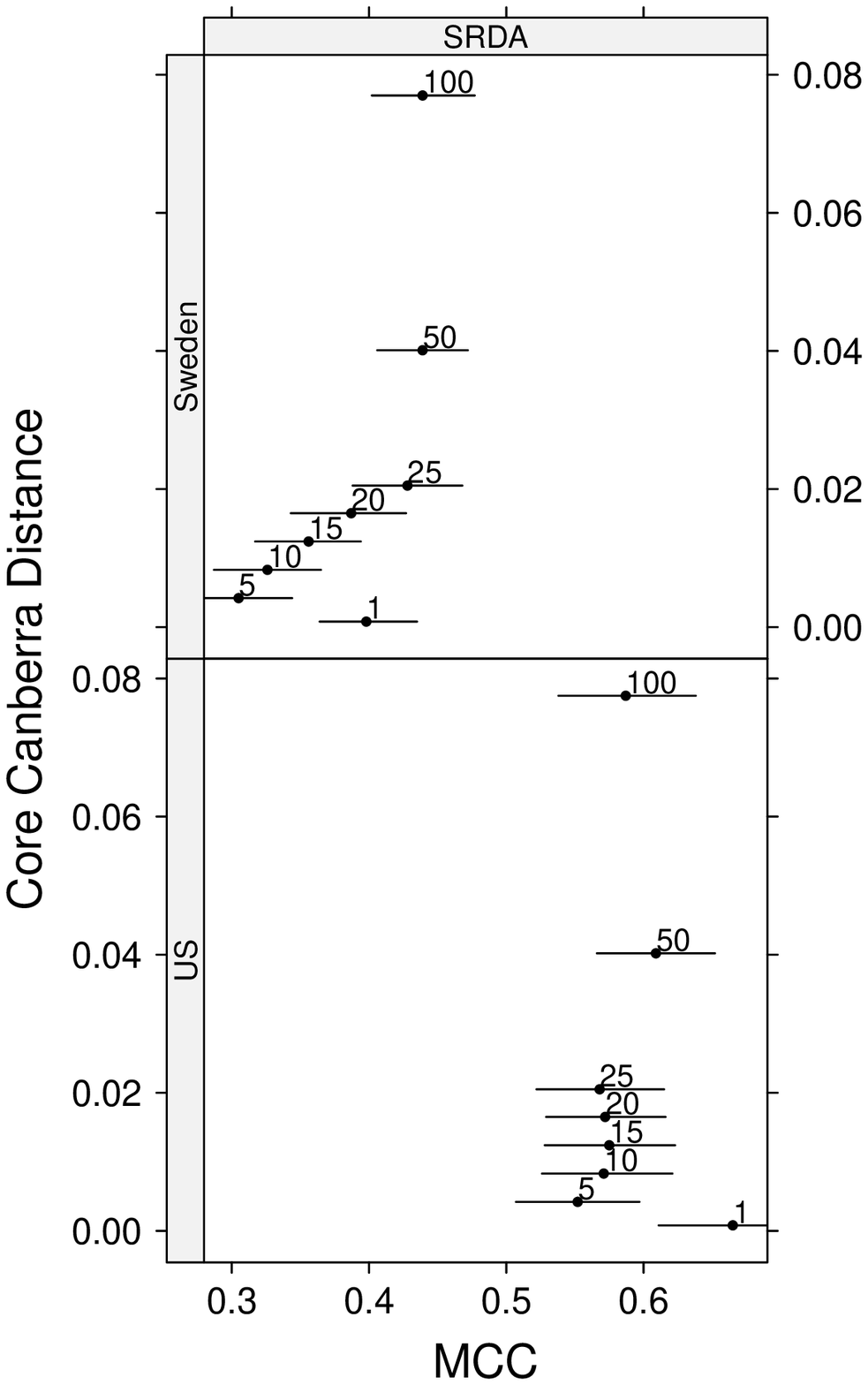}
\textrm{b}
\end{center}
\caption{MCC and Canberra Core values computed by using the SRDA, SVM, 
 and DLDA models on the two Setlur datasets. Each point indicates a model with a fixed number of features, marked above the corresponding CI line.}
\label{fig:setlur-core-both}
\end{figure*}

\begin{figure}[tb]
\begin{center}
\includegraphics[scale=0.44]{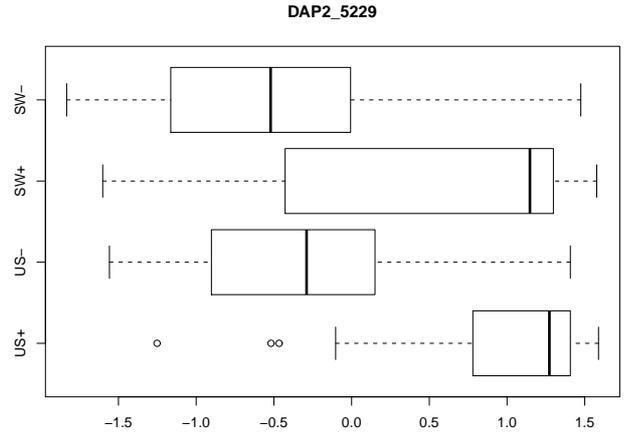}
\end{center}
\caption{Boxplot of the  DAP2\_5229 expression value separately for the two Setlur datasets and the two class labels.}
\label{fig:dap2_5220}
\end{figure}

\begin{figure}[tbh!]
\begin{center}
\includegraphics[scale=0.44]{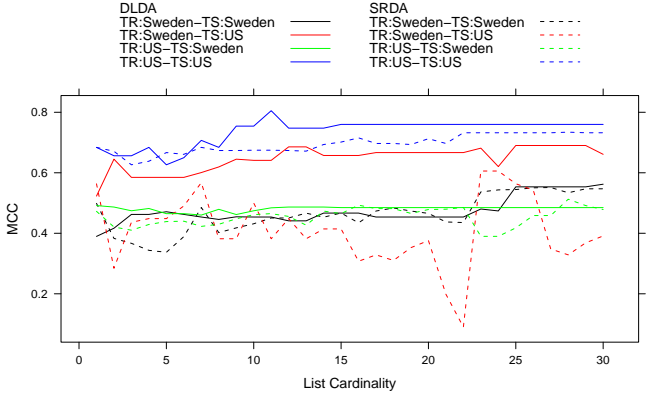}
\end{center}
\caption{MCC for SRDA and DLDA models on increasing number of features extracted from the global list from 1 to 30 on the Setlur data.}
\label{fig:reducing-optimal-list}
\end{figure}

The second application concerns the assessment of the stability of
sets of gene panels derived from profiling tasks, where different
configuration of the learning scheme (e.g. the classifier, or the
ranking algorithm) are compared.  Here the task is how to select a
list of predictive biomarkers and a classifier to predictively discriminate
prostate cancer patients carrying the TMPRSS2-ER gene fusion.
A basic Data Analysis Protocol (DAP for short) is applied to both cohorts of the Setlur dataset, namely a stratified 10$\times$ 5-CV, using three different classifiers: 
Diagonal Linear Discriminant Analysis (DLDA) \cite{pique-regi06block,pique-regi05sequential,bo02new,dudoit02comparison},
linear Support Vector Machines (SVM), and Spectral Regression Discriminant Analysis (SRDA) \cite{cai08srda}. 
A tuning phase through landscaping identified $10^{-3}$ as the optimal value for the SVM regularizer $C$ on both dataset, and $10^3$ and $10^4$ as the two values for the SRDA parameter $\alpha$ respectively on the US and the Sweden cohort (no tuning is needed for the DLDA classifier). 
Furthermore, in the SVM case the dataset is standardized to mean zero and variance one.
The Entropy-based Recursive Feature Elimination (E--RFE, \cite{furlanello03entropy}) 
ranking algorithm is run on the training portion of the cross-validation split 
and classification models with increasing number of best ranked
features are computed on the test part. 
The performances are evaluated at fixed feature set sizes by averaging over the CV replicates the Matthew Correlation
Coefficient (MCC for short, see \cite{baldi00assessing}) Eq. \eqref{eq:mcc}) and the Area Under the ROC Curve (AUC for short) by using the
Wilcoxon-Mann-Whitney formula Eq. \eqref{eq:auc} to extend the measure to binary classifiers. 
In \cite{cortes03auc,calders07efficient,vanderlooy08critical} the equivalence with other formulations is shown: in particular, it is proved that the Wilcoxon-Mann-Whitney formula is an unbiased estimator of the classical AUC.
The two performance metrics adopted have been chosen because they are generally regarded as being two of best measures in describing the confusion matrix (see Tab. \ref{tab:conf}) of true and false positives and negatives by a single number. MCC's range is $[-1,1]$, where $MCC=0$ corresponds to the no-information error rate, which is, for a dataset with $P$ positive samples and $N$ negative samples, equivalent to $\frac{\min\{P,N\}}{P+N}$. MCC=1 is the perfect classification (FP=FN=0), while MCC=-1 denotes the worst possible performance TN=TP=0.
\begin{table}[tb]
\tiny
\begin{tabular*}{\columnwidth}{@{\extracolsep{\fill}}ccccc@{\extracolsep{\fill}}}
\hline
&&& \multicolumn{2}{c}{Actual value} \\
&&& Positive & Negative \\
\hline
Predicted &$\ $& Positive & TP & FP \\
value && Negative & FN & TN \\
\hline
\end{tabular*}
\caption{Confusion matrix for a binary problem; T/F: true/false; TP+FN: all positive samples, TN+FP: all negative samples.}
\label{tab:conf}%
\normalsize
\end{table}
\begin{align}
\label{eq:mcc}
&\textrm{MCC} = \frac{\textrm{TP}\cdot\textrm{TN}-\textrm{FP}\cdot\textrm{FN}}{\sqrt{\left(\textrm{TP}+\textrm{FP}\right)\left(\textrm{TP}+\textrm{FN}\right)\left(\textrm{TN}+\textrm{FP}\right)\left(\textrm{TN}+\textrm{FN}\right)}}\ ,\\
&\textrm{where TN, FP, FN, TP as in Tab. \ref{tab:conf}}\ .\hfill\nonumber
\end{align}
\begin{align}
\label{eq:auc}
&\textrm{AUC} = \frac{\sum_{i=1}^{n_+} \sum_{j=1}^{n_-} I(f(x^+_i) > f(x^-_j)}{n_+ n_-}\ ,\\
&\textrm{where $f$ classifier}, \textrm{$\{x^+_i\}^{n_+}_1$ positive, $\{x^-_j\}^{n_-}_1$ negative.}\hfill\nonumber
\end{align}
In Tabs. \ref{tab:srda-mcc} and \ref{tab:svm} we report the performances on SVM and SRDA on discrete steps of top ranked features ranging from 5 to 6144, with 95\% bootstrap confidence intervals; for comparison purposes we also report AUC values for the SRDA classifier in Tab. \ref{tab:srda-auc}.
For the same values $k$ of the feature set sizes, the Canberra Core Distance is also computed on the top-$k$ ranked lists as produced by the E-RFE algorithm: the stability is also shown in the same tables.
DLDA automatically choses the optimal number of features to use in order to maximize MCC (by tuning the internal parameter $nf$, starting from the default value $nf=0$), thus it is meaningless evaluating this classifier on a different feature set size. 
In particular, DLDA reaches maximal performances with one feature (which is the same for all replicates, DAP2\_5229, leading to a zero stability value): the resulting MCC is 0.26 (CI: (0.18, 0.34)) and 0.16 (CI: 0.12, 0.19) respectively for the US and the Sweden cohort.
As a reference, 5-CV with $9$-NN (which has higher performance than $k=\{5,7,11\}$) has MCC 0.36 on both cohorts with all features.
\begin{table}[tb]
\tiny
\begin{tabular*}{\columnwidth}{@{\extracolsep{\fill}}rrrrrrr@{\extracolsep{\fill}}}
\hline  
& \multicolumn{3}{c}{US} & \multicolumn{3}{c}{Sweden}\\
step & MCC & CI 95\% & Core & MCC & CI 95\% & Core \\
\hline
  1 & 0.00 & (0.00;0.00) & 0.00 & 0.00 & (0.00;0.00) & 0.00 \\
  5 & 0.00 & (0.00;0.00) & 0.00 & 0.00 & (0.00;0.00) & 0.00 \\
 10 & 0.00 & (0.00;0.00) & 0.01 & 0.00 & (0.00;0.00) & 0.01 \\
 15 & 0.00 & (0.00;0.00) & 0.01 & 0.00 & (0.00;0.00) & 0.01 \\
 20 & 0.00 & (0.00;0.00) & 0.02 & 0.00 & (0.00;0.00) & 0.02 \\
 25 & 0.00 & (0.00;0.00) & 0.02 & 0.00 & (0.00;0.00) & 0.02 \\
 50 & 0.00 & (0.00;0.00) & 0.04 & 0.00 & (0.00;0.00) & 0.04 \\
100 & 0.00 & (0.00;0.00) & 0.08 & 0.00 & (0.00;0.00) & 0.08 \\
1000 & 0.51 & (0.47;0.56) & 0.52 & 0.08 & (0.05;0.12) & 0.52 \\
5000 & 0.53 & (0.49;0.58) & 0.88 & 0.23 & (0.20;0.27) & 0.91 \\
6144 & 0.53 & (0.49;0.58) & 0.59 & 0.24 & (0.20;0.27) & 0.62 \\
\hline
\end{tabular*}
\caption{MCC and Core Canberra values for the two Setlur datasets for SVM classifiers.}
\label{tab:svm}%
\normalsize
\end{table}
\begin{table}[tb]
\tiny
\begin{tabular*}{\columnwidth}{@{\extracolsep{\fill}}rrrrrrr@{\extracolsep{\fill}}}
\hline
& \multicolumn{3}{c}{US} & \multicolumn{3}{c}{Sweden}\\
step & MCC & CI 95\% & Core & MCC & CI 95\% & Core \\
\hline    
1 & 0.67 & (0.61;0.72) & 0.00 & 0.40 & (0.36;0.43) & 0.00 \\
   5 & 0.55 & (0.51;0.60) & 0.00 & 0.30 & (0.26;0.34) & 0.00 \\
  10 & 0.57 & (0.53;0.62) & 0.01 & 0.33 & (0.29;0.36) & 0.01 \\
  15 & 0.57 & (0.53;0.62) & 0.01 & 0.36 & (0.32;0.39) & 0.01 \\
  20 & 0.57 & (0.53;0.62) & 0.02 & 0.39 & (0.34;0.43) & 0.02 \\
  25 & 0.57 & (0.52;0.61) & 0.02 & 0.43 & (0.39;0.47) & 0.02 \\
  50 & 0.61 & (0.57;0.65) & 0.04 & 0.44 & (0.41;0.47) & 0.04 \\
 100 & 0.59 & (0.54;0.64) & 0.08 & 0.44 & (0.40;0.48) & 0.08 \\
1000 & 0.50 & (0.45;0.55) & 0.52 & 0.47 & (0.43;0.50) & 0.51 \\
5000 & 0.51 & (0.46;0.56) & 0.89 & 0.46 & (0.43;0.50) & 0.84 \\
6144 & 0.51 & (0.46;0.56) & 0.60 & 0.46 & (0.42;0.49) & 0.52 \\
\hline
\end{tabular*}
\caption{MCC and Core Canberra values for the two Setlur datasets for SRDA classifiers.}
\label{tab:srda-mcc}%
\normalsize
\end{table}
\begin{table}[tb]%
\tiny
\begin{tabular*}{\columnwidth}{@{\extracolsep{\fill}}rrrrr@{\extracolsep{\fill}}}
\hline  
& \multicolumn{2}{c}{US} & \multicolumn{2}{c}{Sweden}\\
 step & AUC & CI 95\% & AUC & CI 95\% \\
\hline
1 & 0.87 & (0.84;0.89) & 0.79 & (0.77;0.80)\\
  5 & 0.83 & (0.81;0.85) & 0.79 & (0.77;0.80)\\
 10 & 0.86 & (0.84;0.88) & 0.80 & (0.79;0.82)\\
 15 & 0.88 & (0.86;0.89) & 0.82 & (0.81;0.83)\\
 20 & 0.88 & (0.86;0.90) & 0.83 & (0.81;0.84)\\
 25 & 0.89 & (0.87;0.91) & 0.84 & (0.82;0.85)\\
 50 & 0.90 & (0.89;0.92) & 0.84 & (0.83;0.86)\\
100 & 0.90 & (0.88;0.92) & 0.85 & (0.84;0.86)\\
1000 & 0.86 & (0.85;0.88) & 0.83 & (0.81;0.84)\\
5000 & 0.86 & (0.84;0.88) & 0.82 & (0.81;0.84)\\
6144 & 0.86 & (0.84;0.88) & 0.82 & (0.81;0.84)\\
\hline
\end{tabular*}
\caption{AUC values for the two Setlur datasets for SRDA classifiers.}
\label{tab:srda-auc}%
\normalsize
\end{table}
All results are displayed in the performance/stability plots of Fig. \ref{fig:setlur-core-both}b. 
These plots can be used as a diagnostic for model selection to detect a possible choice for the optimal model as a reasonable compromise between good performances (towards the rightmost part of the graph) and good stability (towards the bottom of the graph). 
For instance, in the case shown we decide to use SRDA as the better classifier, using 25 features on the Sweden cohort and 10 on the US cohort: looking at the zoomed graph in Fig. \ref{fig:setlur-core-both}b, if we suppose that the points are describing an ideal Pareto front, the two chosen models are the closest to the bottom right corner of the plots. 
The corresponding Borda optimal lists for SRDA models on the two Setlur datasets are detailed in Tab. \ref{tab:optimal-lists}: 5 probes are common to the two list, and, in particular, the top ranked probe is the same.
\begin{table}[!t]
\tiny
\begin{tabular*}{\columnwidth}{@{\extracolsep{\fill}}lrlr@{\extracolsep{\fill}}}
\hline    
\multicolumn{1}{c}{Sweden} & \multicolumn{1}{c}{Ranking}  & \multicolumn{1}{c}{US} & \multicolumn{1}{c}{Ranking}\\
& \multicolumn{1}{c}{in US} &  & \multicolumn{1}{c}{in Sweden}\\
\hline    
\textbf{\textit{DAP2\_5229}}  & \textbf{\textit{1}} &    \textbf{\textit{DAP2\_5229}} & \textbf{\textit{1}}   \\
   \textbf{\textit{DAP1\_2857}}  & \textbf{\textit{5}} &    \textbf{DAP1\_5091} & \textbf{18}  \\
   \textbf{\textit{DAP4\_2051}}  & \textbf{\textit{3}} &    \textbf{\textit{DAP4\_2051}} & \textbf{\textit{3}}   \\
   \textit{DAP1\_1759}           & \textit{13} &   DAP2\_1680          & 51  \\
   DAP1\_2222           & 19 &   \textbf{\textit{DAP1\_2857}} & \textbf{\textit{2}}   \\
   \textit{DAP4\_0822}           & \textit{44} &   \textbf{DAP3\_0905} & \textbf{8}   \\
   \textit{DAP2\_0361}           & \textit{403} &  DAP2\_5769          & \textit{77}  \\
   \textbf{DAP3\_0905}  & \textbf{6} &    DAP4\_2271          & 36  \\
   \textit{DAP2\_5076}           & \textit{24} &   \textit{DAP4\_3958}          & \textit{44}  \\
   \textit{DAP3\_2016}           & \textit{16} &   DAP4\_2442          & 2734\\
   DAP4\_4217           & 497 &  \\
   \textit{DAP2\_0721}           & \textit{421} &  \\
   \textit{DAP4\_1360}           & \textit{18} &   \\
   \textit{DAP3\_1617}           & \textit{15} &   \\
   DAP1\_5829           & 529 &  \\
   \textit{DAP3\_6085}           & \textit{12} &   \\
   DAP4\_2180           & 26 &   \\
   \textbf{DAP1\_5091}  & \textbf{2} &    \\
   DAP1\_2043           & 1989 & \\
   \textit{DAP4\_2027}           & \textit{2227} & \\
   \textit{DAP4\_1375}           & \textit{145} &  \\
   DAP4\_5930           & 3455 & \\
   \textit{DAP4\_4205}           & \textit{25} &   \\
   DAP1\_4950           & 166 &  \\
   \textit{DAP4\_1577}           & \textit{283} &  \\
\hline  
\end{tabular*}
\caption{Borda optimal lists for SRDA models on the two Setlur datasets. In boldface, probes common to the two optimal lists. In italic, probes included in the 87-gene signature of the original paper \cite{setlur08estrogen}. 17 probes out of 30 are common to the 87-gene signature in \cite{setlur08estrogen}}
\label{tab:optimal-lists}
\normalsize
\end{table}
In Tab. \ref{tab:mcc-optimal-models} we list the MCC obtained by applying the SRDA and DLDA models on the two Setlur cohorts (exchanging their role as training and test set) by using the two optimal Borda lists.
\begin{table}[tb]
\tiny
\begin{tabular*}{\columnwidth}{@{\extracolsep{\fill}}lllrr@{\extracolsep{\fill}}}
\hline    
Borda & Training & Test & SRDA & DLDA\\
\hline    
US & US & Sweden         & 0.39  & 0.44\\
Sweden & Sweden & US     & 0.42  & 0.48\\    
\hline    
US & Sweden & US         & 0.48  & 0.63\\
Sweden & US & Sweden     & 0.51  & 0.45\\
\hline    
US & Sweden & Sweden     & 0.39  & 0.45\\
Sweden & US & US         & 0.69  & 0.71\\
\hline    
US & US & US             & 0.71  & 0.78\\
Sweden & Sweden & Sweden & 0.55  & 0.52\\
\hline
\end{tabular*}
\caption{Setlur dataset. MCC values for SRDA and DLDA optimal models.}
\label{tab:mcc-optimal-models}
\normalsize
\end{table}
The probe DAP2\_5229 probe seems to have a relevant discriminative and predictive importance, as shown by the classwise boxplots on the two cohorts of Fig. \ref{fig:dap2_5220}.
As detailed in GEO \url{http://www.ncbi.nlm.nih.gov/geo} and in NCBI Nucleotide DB \url{http://www.ncbi.nlm.nih.gov/nuccore/}, its RefSeq ID is NM\_004449, 
whose functional description is reported as ``v-ets erythroblastosis virus E26 oncogene homolog (avian) (ERG), transcript variant 2, mRNA'' (information updated on 28 June 2009).
In Tab. \ref{tab:mcc-optimal-models-1f+g} we show the performances obtained by a SRDA and a DLDA model with the sole feature DAP2\_5229 on all combinations of US and Sweden cohort as training and test set.
If we consider as the global optimal list the list of all 30 distinct features given as the union of the Borda list in Tab.~\ref{tab:optimal-lists}, we get for SRDA and DLDA models the performances listed in Tab.~\ref{tab:mcc-optimal-models-1f+g}. 
\begin{table}[tb]
\tiny
\begin{tabular*}{\columnwidth}{@{\extracolsep{\fill}}llrrrr@{\extracolsep{\fill}}}
\hline            
&      & \multicolumn{2}{c}{SRDA} & \multicolumn{2}{c}{DLDA}\\ 
   Training & Test & DAP2 & global & DAP2 & global \\
    &  & 5229 & optimal & 5229 & optimal\\
\hline    
US & Sweden & 0.47     & 0.47 & 0.49 & 0.48\\
   Sweden & US & 0.56     & 0.39 & 0.52 & 0.66\\
   Sweden & Sweden & 0.50 & 0.55 & 0.39 & 0.56\\
   US & US & 0.68         & 0.73 & 0.68 & 0.76\\
\hline  
\end{tabular*}
\caption{MCC values for SRDA and DLDA models with the only feature DAP2\_5229 and with the global optimal list.}
\label{tab:mcc-optimal-models-1f+g}
\normalsize
\end{table}
To check the consistency of the retrieved global list, we run a permutation test: we randomly extract 30 features out of the original 6144 features and we use as the p-value the number of times the obtained performances (DLDA models) are better than those obtained with the global optimal list, divided by the total number $10^4$ of experiments. 
The resulting p-values are less than $10^-3$ for all four combinations of using the two cohorts as training and test set, thus obtaining a reasonable significance of the global optimal list. 
Nevertheless if the same permutation test is run with the feature DAP2\_5229 always occurring in the chosen random feature sets, the results are very different: namely, the p-value results about $0.1$, thus indicating a small statistical significance of the obtained global list. 
These tests seems to indicate that the occurrence of DAP2\_5229 plays a key role in finding a correct predictive signature. 
We then performed a further experiment to detect the predictive power of the global optimal list as a function of its length. 
We order the global list keeping DAP2\_5229, DAP4\_2051, DAP1\_2857, DAP3\_0905, and DAP1\_5091 as the first five probes and compute the performances of a DLDA model by increasing the number of features extracted from the global list from 1 to 30. 
The result is shown in Fig.~\ref{fig:reducing-optimal-list}: for many of the displayed models a reduced optimal list of about 10-12 features is sufficient to get almost optimal predictive performances. 
A permutation test on 12 features (with DAP2\_5229 kept as the top probe) gives a p-value of $10^-2$.
A final note: our results show a slightly better AUC (in training) than the one found by the authors of the original paper \cite{setlur08estrogen}, both in the Sweden and in the US cohort. 
Moreover, 17 out of 30 genes included in the global optimal list are member of the 87-gene signature shown in the original paper.

\subsection{Comparing}
\begin{figure*}[tb]
\begin{center}
\textbf{(a)}
\includegraphics[width=0.45\textwidth]{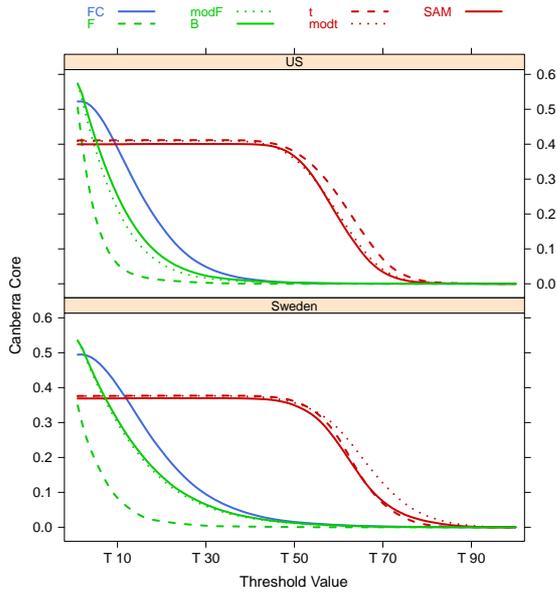}
\includegraphics[width=0.45\textwidth]{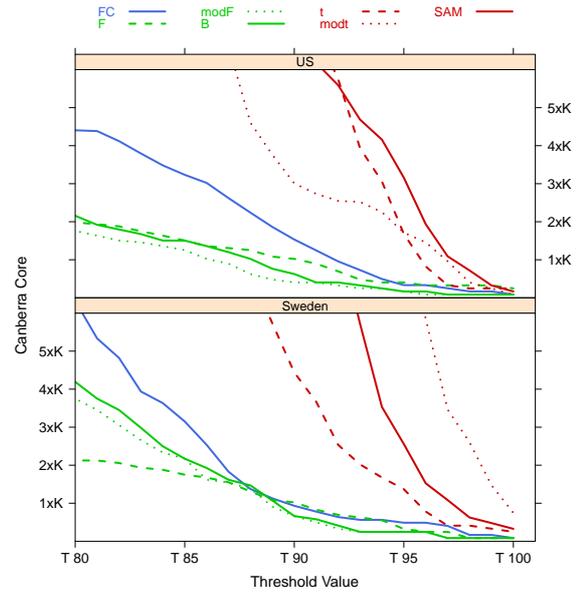}
\textbf{(b)}
\end{center}
\caption{\textbf{(a)}: Canberra core evaluated on the Setlur dataset on B=100 repeated filtering experiments on 90\% of the data. \textbf{(b)} Zoom on the 80\%-100\% threshold zone. $K=10^5$}
\label{fig:filtering-setlur-both}
\end{figure*}

The seven filtering algorithms of the previous subsection are also applied to the Setlur dataset, by using 100 resampling on 90\% of the data on both the US and Sweden cohort separately. 
The Canberra Core values of the lists at different values of the filtering thresholds are shown in Fig. \ref{fig:filtering-setlur-both}, together with a zoom on the stricter constraints area: the plots confirm the different behaviour of the groups $\{t,\textrm{mod-}t,\textrm{SAM}\}$ and $\{F,\textrm{mod-}F,\textrm{B}\}$ and of the singleton $FC$ in both cases.
By considering a cutting threshold of the 75\% of the maximal value, we retrieve 14 sets of ranked partial lists, from which 14 Borda optimal lists are computed.
In Tab. \ref{tab:borda-filtering-list-len} we list the lengths of the Borda lists for each filtering method and cohort.
\begin{table}[b]
\tiny

\begin{tabular*}{\columnwidth}{@{\extracolsep{\fill}}rrrrrrrr@{\extracolsep{\fill}}}
\hline      
& F & FC & modF & modt & t & B & SAM \\
\hline      
Sweden &   1 &  17 &  25 & 759 & 326 &  28 & 366 \\
US &   1 &   3 &   6 & 208 & 367 &   7 & 149 \\
\hline    
\end{tabular*}
\caption{Length of the Borda lists for different filtering methods at 75\% threshold on the Setlur dataset.}
\label{tab:borda-filtering-list-len}
\end{table}

\begin{table*}[bt]
\tiny
\begin{tabular*}{\textwidth}{@{\extracolsep{\fill}}lrrrrrrrrrrrrrrrrr@{\extracolsep{\fill}}}
\hline  
& \textit{F} & \textit{FC} & \textit{modF} & \textit{modt} & \textit{t} & \textit{B} & \textit{SAM} & \textit{SRDA} & F & FC & modF & modt & t & B & SAM & SRDA \\
\hline         
 \textit{F}    & $\blacksquare$  & 0.007 & 0.011 & 0.230 & 0.115 & 0.012 & 0.127 & 0.010 & 0.000 & 0.001 & 0.003 & 0.077 & 0.127 & 0.003 & 0.057 & 0.004 \\
 \textit{FC}   & \textbf{122} & $\blacksquare$ & 0.016 & 0.231 & 0.116 & 0.018 & 0.128 & 0.017 & 0.007 & 0.008 & 0.009 & 0.084 & 0.134 & 0.009 & 0.064 & 0.010 \\
 \textit{modF} & \textbf{69} & \textbf{129} & $\blacksquare$ & 0.228 & 0.114 & 0.002 & 0.126 & 0.021 & 0.011 & 0.012 & 0.013 & 0.087 & 0.136 & 0.013 & 0.067 & 0.014 \\
 \textit{modt} & \textbf{7324} & \textbf{7337} & \textbf{7307} & $\blacksquare$ & 0.165 & 0.228 & 0.163 & 0.239 & 0.230 & 0.231 & 0.232 & 0.303 & 0.352 & 0.232 & 0.283 & 0.234 \\
 \textit{t}    & \textbf{2418} & \textbf{2441} & \textbf{2401} & \textbf{7379} & $\blacksquare$ & 0.115 & 0.108 & 0.125 & 0.115 & 0.116 & 0.117 & 0.192 & 0.244 & 0.118 & 0.173 & 0.119 \\
 \textit{B}    & \textbf{73} & \textbf{132} & \textbf{75} & \textbf{7308} & \textbf{2402} & $\blacksquare$ & 0.127 & 0.022 & 0.012 & 0.013 & 0.014 & 0.088 & 0.138 & 0.014 & 0.068 & 0.016 \\
 \textit{SAM}  & \textbf{3925} & \textbf{3924} & \textbf{3912} & \textbf{7287} & \textbf{4084} & \textbf{3914} & $\blacksquare$ & 0.136 & 0.127 & 0.128 & 0.129 & 0.201 & 0.250 & 0.130 & 0.181 & 0.131 \\
 \textit{SRDA} & \textbf{998} & \textbf{1116} & \textbf{1067} & \textbf{8326} & \textbf{3423} & \textbf{1071} & \textbf{4916} & $\blacksquare$ & 0.010 & 0.009 & 0.012 & 0.084 & 0.133 & 0.012 & 0.062 & 0.011 \\
 F    & \textbf{19} & \textbf{115} & \textbf{63} & \textbf{7317} & \textbf{2412} & \textbf{66} & \textbf{3919} & \textbf{1004} & $\blacksquare$ & 0.001 & 0.003 & 0.077 & 0.127 & 0.003 & 0.057 & 0.004 \\
 FC   & \textbf{51} & \textbf{159} & \textbf{106} & \textbf{7360} & \textbf{2455} & \textbf{110} & \textbf{3962} & \textbf{976} & \textbf{55} & $\blacksquare$ & 0.004 & 0.077 & 0.127 & 0.004 & 0.057 & 0.003 \\
 modF & \textbf{52} & \textbf{111} & \textbf{59} & \textbf{7313} & \textbf{2408} & \textbf{63} & \textbf{3915} & \textbf{1049} & \textbf{45} & \textbf{88} & $\blacksquare$ & 0.077 & 0.127 & 0.001 & 0.057 & 0.005 \\
 modt & \textbf{1124} & \textbf{1216} & \textbf{1162} & \textbf{8393} & \textbf{3478} & \textbf{1165} & \textbf{4990} & \textbf{2032} & \textbf{1124} & \textbf{1123} & \textbf{1126} & $\blacksquare$ & 0.066 & 0.078 & 0.052 & 0.078 \\
 t    & \textbf{2194} & \textbf{2284} & \textbf{2229} & \textbf{9449} & \textbf{4535} & \textbf{2233} & \textbf{6048} & \textbf{3070} & \textbf{2194} & \textbf{2195} & \textbf{2195} & \textbf{2081} & $\blacksquare$ & 0.128 & 0.094 & 0.128 \\
 B    & \textbf{60} & \textbf{120} & \textbf{67} & \textbf{7321} & \textbf{2416} & \textbf{71} & \textbf{3923} & \textbf{1057} & \textbf{53} & \textbf{97} & \textbf{29} & \textbf{1126} & \textbf{2196} & $\blacksquare$ & 0.058 & 0.006 \\
 SAM  & \textbf{1002} & \textbf{1095} & \textbf{1041} & \textbf{8283} & \textbf{3371} & \textbf{1045} & \textbf{4879} & \textbf{1843} & \textbf{1003} & \textbf{997} & \textbf{1004} & \textbf{1188} & \textbf{2190} & \textbf{1004} & $\blacksquare$ & 0.057 \\
 SRDA & \textbf{385} & \textbf{504} & \textbf{455} & \textbf{7711} & \textbf{2806} & \textbf{459} & \textbf{4311} & \textbf{1015} & \textbf{392} & \textbf{370} & \textbf{436} & \textbf{1406} & \textbf{2470} & \textbf{445} & \textbf{1241} & $\blacksquare$ \\
\hline
\end{tabular*}
\caption{Distances between Borda optimal lists (upper triangular matrix) and between all partial lists (lower triangular matrix, $\times 10^{5}$) for filtering methods (75\% threshold) and SRDA models. Rows and columns 1-8 (\textit{Italic}): Sweden cohort; rows and columns 9-16: US cohort.}
\label{tab:borda-dist-all}
\normalsize
\end{table*}

\begin{figure*}[bth!]
\begin{center}
\textbf{(a)}
\includegraphics[width=0.45\textwidth]{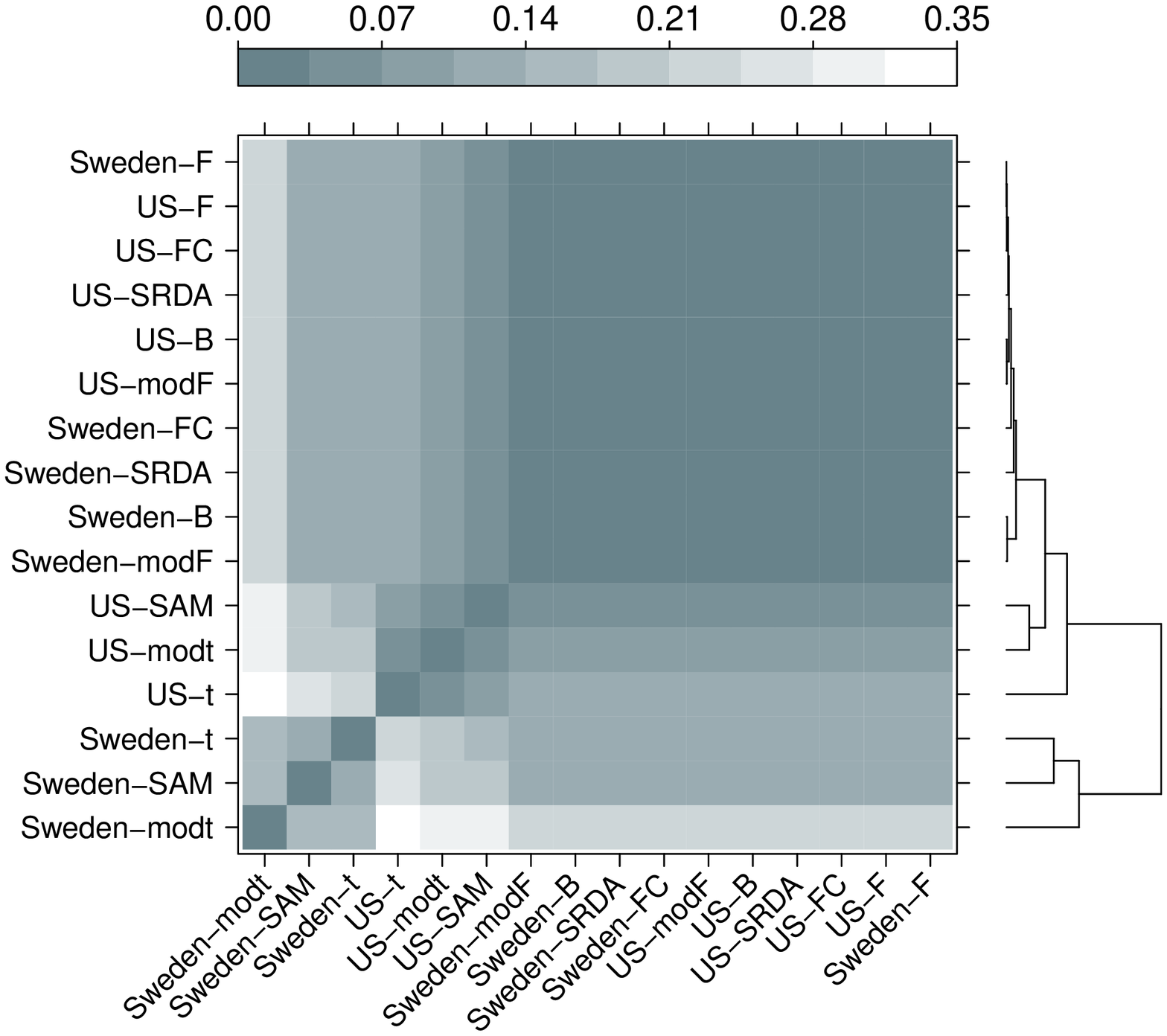} 
\includegraphics[width=0.45\textwidth]{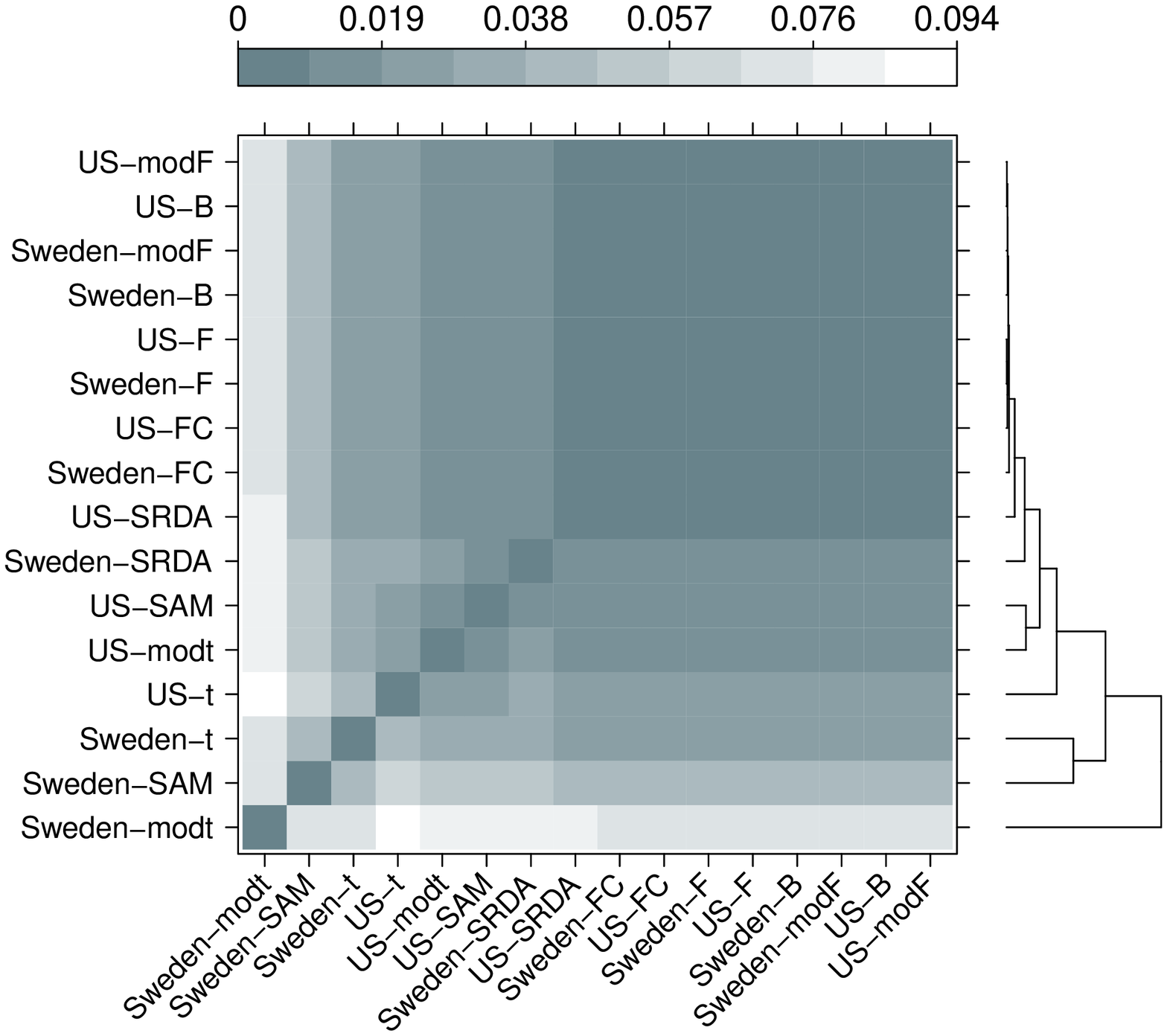} 
\textbf{(b)}
\end{center}
\caption{Levelplot of the distances computed on the lists produced by filtering methods (75\% threshold) and SRDA models, where the Canberra Distance is computed on \textbf{(a)} their Borda lists; \textbf{(b}) their whole list sets. }
\label{fig:dist-all}
\end{figure*}

\begin{figure*}[tbh!]
\begin{center}
\textbf{(a)}
\includegraphics[width=0.45\textwidth]{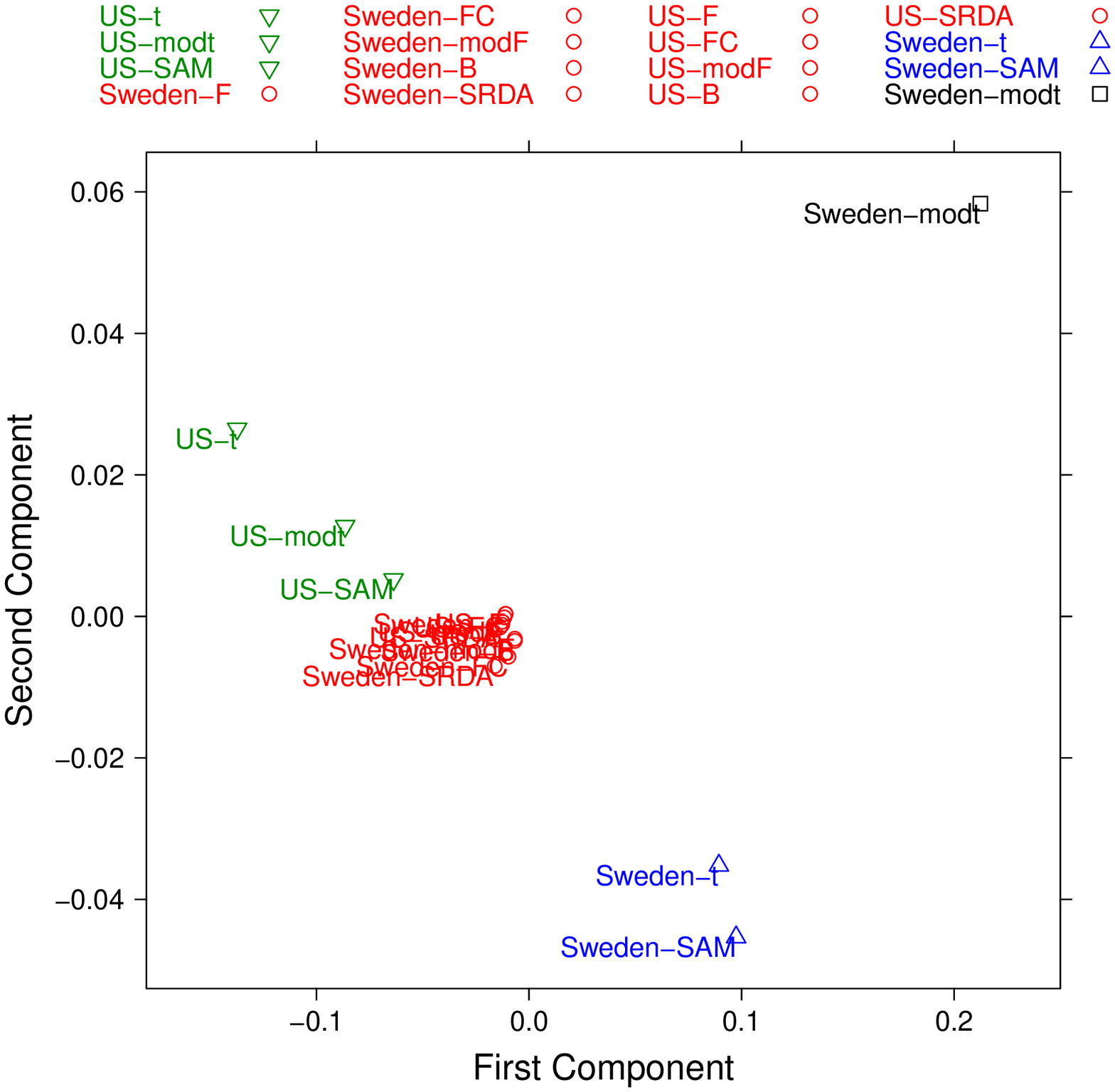} 
\includegraphics[width=0.45\textwidth]{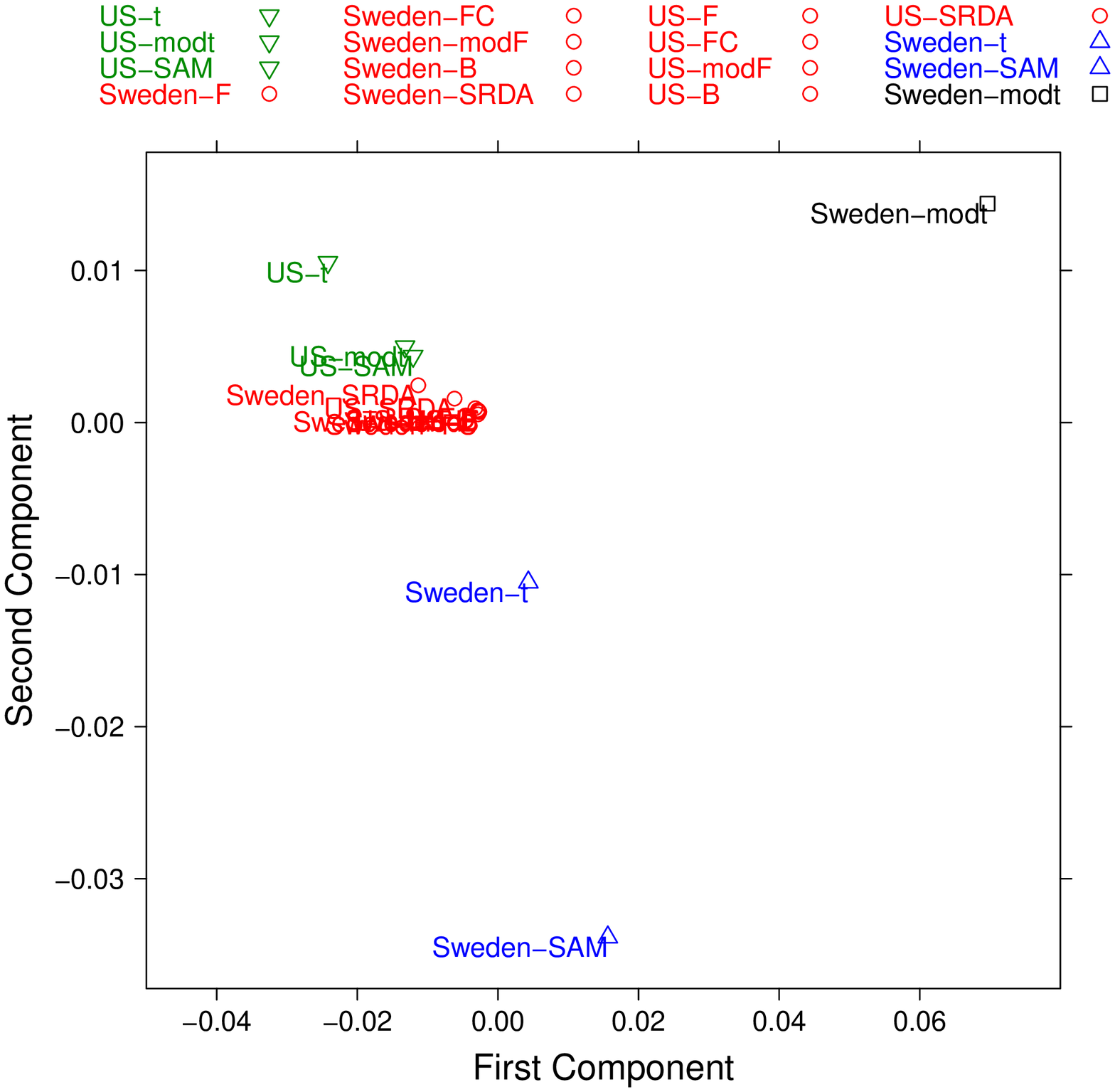} 
\textbf{(b)}
\end{center}
\label{fig:mds-all-list}
\caption{Multidimensional Scaling (MDS) on two components computed on the lists produced by filtering methods (75\% threshold) and SRDA models, where the Canberra distance is computed on \textbf{(a)} their Borda lists; \textbf{(b)} their whole list sets. }
\end{figure*}

As a first rough set-theoretical comparison, we list in Tab. \ref{tab:borda-filtering-lists} the probes common to more than three filtering methods. We note that only three probes are also appearing in the corresponding SRDA Borda list.
\begin{table}[tb]
\tiny
\begin{tabular*}{\columnwidth}{@{\extracolsep{\fill}}lclc@{\extracolsep{\fill}}}
\hline      
\multicolumn{2}{c}{Sweden}&\multicolumn{2}{c}{US}\\
gene & extractions &      gene & extractions\\
\hline      
DAP2\_1768 & 6 & DAP2\_4092 & 5 \\
DAP1\_1949 & 5 & DAP2\_5047 & 5 \\
DAP1\_4198 & 5 & \textbf{DAP2\_5229} & \textbf{5} \\
DAP1\_5095 & 5 & \textbf{DAP4\_2442} & \textbf{5} \\
DAP2\_1037 & 5 & \textbf{DAP4\_2051} & \textbf{4} \\
DAP2\_1151 & 5 &  &  \\
DAP2\_3790 & 5 &  &  \\
DAP2\_3896 & 5 &  &  \\
DAP2\_5650 & 5 &  &  \\
DAP3\_2164 & 5 &  &  \\
DAP3\_4283 & 5 &  &  \\
DAP3\_5834 & 5 &  &  \\
DAP4\_1974 & 5 &  &  \\
DAP4\_2316 & 5 &  &  \\
DAP4\_4178 & 5 &  &  \\
13 genes   & 4 &  &  \\
\hline    
\end{tabular*}
\label{tab:borda-filtering-lists}
\caption{List of probes common to more than three filtering methods}
\normalsize
\end{table}
In order to get a more refined indicator of similarity, we also compute the Core Canberra Distances between all Borda optimal lists and between all 75\%-threshold partial lists for filtering methods, together with the corresponding partial and Borda lists for the SRDA models: all results are reported in Tab. \ref{tab:borda-dist-all}
By using the Core distances, we draw two levelplots (for both distances on Borda lists and on the whole partial lists sets, computing also a hierarchical cluster with average linkage and representing also the corresponding dendrogram in Fig. \ref{fig:dist-all}.
A further graphical representation of the computed distances has been obtained by using a Multidimensional Scaling (MDS) on two components, as shown in Fig. \ref{fig:mds-all-list}.
On both cases, a few facts emerges: on both cohorts, the results on the Borda lists and on the whole sets of lists are similar, indicating that the Borda method is a good way to incorporate information into a single list; the behaviour grouping detected in the previous subsection is essentially confirmed here. Moreover, the two cohort are quite different, while the lists coming from the profiling experiments are not deeply different from those emerging by the filtering methods.



\section{DISCUSSION}
In \cite{shi09tale}, a correlation between signature congruency and model performance in MAQC-II \cite{maqc09maqcII} has been detected both in training and validation sets: the more similar the signatures, the better the average predictions.

The range of possible applications is clearly not limited to the example shown in the present work: in \cite{boulesteix09stability,ein-dor06thousands,boutros09prognostic,lau07three-gene} the importance of defining indicators for assessing ranked lists variability is discussed.
At least two more applications are worth mentioning, metanalysis studies to investigate relationship between stability and classification accuracy (see for instance \cite{maqc09maqcII}) and analysis of lists produced by methods of gene lists enrichment.

As a final consideration, the described method may become an essential tool towards a theoretical improvement in the workfield, that is, the construction of a stability theory for feature selection, for instance in the leave-one-out stability theory already developed for classifiers \cite{mukherjee06learning}, \cite{bousquet02stability}. 
Attempts in this direction has been recently carried out \cite{kalousis05stability}, \cite{kuncheva07astability}, \cite{zhang07amethod}, \cite{krizek07improving}, \cite{xiao07quantification}, but so far no general framework has been structured yet.
%


\section{CONCLUSIONS}
In the present work, an effective method for comparing heterogeneous ranked lists coming from different experiments is shown. The algorithm is designed within the framework of the theory of metric methods for permutation groups. The introduced metric can be used in different contexts and for different purposes in many aspects of computational biology. A few examples of use are shown in the final section of the paper.

\section{FUNDING}
European Union (FP7 project HiperDart). 

\section{ACKNOWLEDGMENTS}
The authors would like to thank Davide Albanese for the implementation within the \texttt{mlpy} package and Silvano Paoli for his support while running computation on the FBK HPC facility.
%

\newpage
\bibliographystyle{NAR}
\bibliography{PartialLists}
\end{document}